# Learning Projection-Aware 360-Degree Image Rectification via Dual-Projection Fusion

Yuhao Shan, Qianyi Yuan, Jingguo Liu, Shigang Li, Jianfeng Li, Tong Chen

***Abstract*— Panoramic cameras provide a 360° field of view and are widely used in panoramic vision, immersive visual computing, and robotic perception. However, changes in camera orientation can produce non-upright panoramas, introducing geometric variations that can complicate downstream visual analysis. Existing vision-based rectification methods usually operate within a single projection domain, limiting their ability to jointly exploit local geometric structures and global contextual information. To address this, we formulate 360° image rectification as a projection-aware representation learning problem and propose a dual-projection framework for upright panoramic rectification. A convolutional neural network branch captures local geometric structures from equirectangular projection (ERP) inputs, while a vision transformer branch models global contextual cues from cubemap projections. Cross-projection feature transformation and multi-level feature fusion enable effective interaction between these complementary representations. The learned representation supports collaborative inclination estimation and upright panorama generation, with the two tasks providing complementary geometric and appearance supervision. Experiments on SUN360 and M3D show consistent improvements over existing methods, achieving accuracies within a 1° error threshold of 65.9% and 85.2% and Fréchet Inception Distance scores of 5.87 and 3.26, respectively. Ablation studies verify the contributions of dual-projection representation, cross-projection feature transformation and fusion, and collaborative multi-task learning. The proposed framework provides a projection-aware visual computing approach for panoramic rectification. Code, pretrained models, and training/testing scripts are available at https://github.com/YuhaoShine/DualProjectionFusion.**

## I. INTRODUCTION

PANORAMIC cameras, with their ability to capture a 360-degree field of view, provide wide visual coverage for panoramic vision and robotic perception. As illustrated in Fig. 1(a), panoramic images acquired by such cameras can be represented on the viewing sphere and further mapped to a two-dimensional equirectangular projection (ERP). Their wide field of view is particularly useful for applications such as situation awareness, simultaneous localization and mapping (SLAM), and 3D reconstruction [1–8]. Moreover, panoramic imaging is also widely used in immersive visual computing, including virtual reality (VR) and augmented reality (AR).

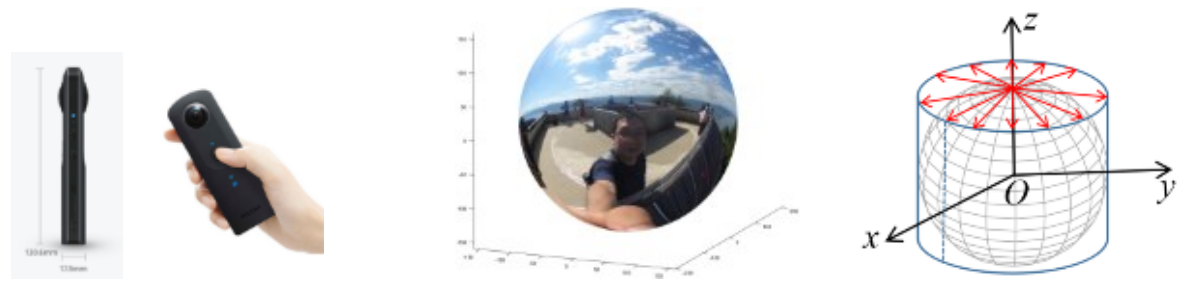

(a). RICOH panoramic camera [9], spherical representation, and ERP.

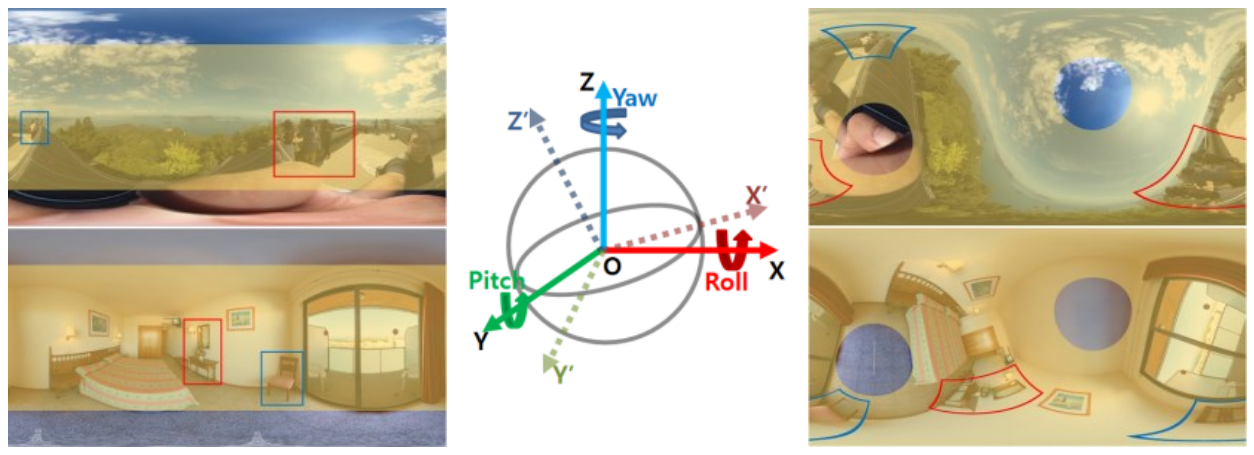

(b). Upright and non-upright ERP images under camera inclination

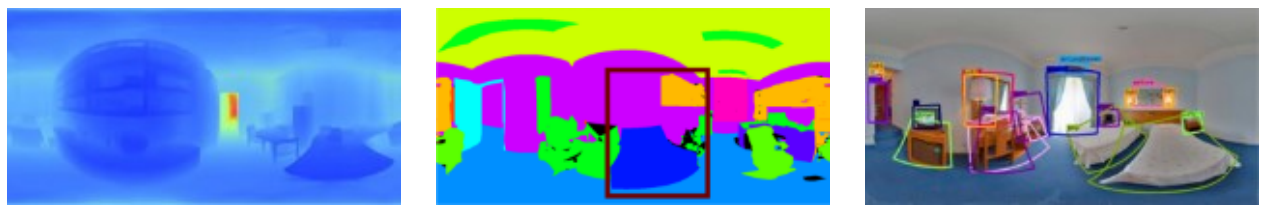

(c). Representative panoramic vision tasks based on upright ERP: depth estimation [10], semantic segmentation [11], and object detection [12].

**Fig. 1.** Overview of panoramic imaging, non-upright ERP formation, and representative panoramic vision tasks.

The ERP is one of the most widely used 2D representations in panoramic vision because it preserves the full 360-degree field of view in a continuous image domain. However, when the panoramic camera deviates from an upright orientation during acquisition, the resulting ERP image becomes non-upright, as illustrated in Fig. 1(b), where the highlighted regions containing perceptually important content, such as tourists and tables/chairs, remain relatively well preserved in the upright ERP but exhibit pronounced geometric distortion in the non-upright case. Such camera inclination introduces additional geometric variations and may increase the difficulty of subsequent panoramic vision tasks, since many existing methods are developed and evaluated using approximately upright ERP inputs [10–13]. Representative examples, including panoramic depth estimation, semantic segmentation, and object detection, are shown in Fig. 1(c). Previous upright-adjustment studies have further demonstrated the practical value of recovering upright panoramic representations in application-oriented scenarios, including real-world observation, pedestrian detection, and depth estimation [14–16]. Therefore, maintaining a consistent upright orientation is

This research was funded by the Fundamental Research Funds for the Central Universities (Grant No. SWU120083), Science and Technology Innovation Key R&D Program of Chongqing (CSTB2023TIAD-STX0037), the National Natural Science Foundation of China (Grant No. 6267340). (Corresponding author: Jianfeng Li, Tong Chen).

Yuhao Shan, Qianyi Yuan, Jingguo Liu, Jianfeng Li, Tong Chen are with the college of electronic and information engineering, Southwest University, Chongqing 400700, China (e-mail: popqlee@swu.edu.cn, c_tong@swu.edu.cn); Shigang Li is with the college of information science, Hiroshima City University, Hiroshima 7313164, Japan.

beneficial for reliable 360-degree visual computing, including robotic perception.

In practical applications, the orientation of a panoramic camera may vary during acquisition. This issue can be particularly relevant on mobile platforms, including robotic systems, where camera orientation changes with platform motion. Although gimbal-based stabilization systems can compensate for such orientation changes, they typically require additional hardware and inertial sensing and may still be affected by sensing or environmental disturbances. In such cases, vision-based methods provide a complementary solution by exploiting geometric cues directly from panoramic images, providing an image-based means of estimating and correcting camera inclination. Existing vision-based approaches can be broadly categorized into two paradigms: **conventional geometry-based methods** and **deep learning-based methods**.

**Conventional geometry-based methods**: When the panoramic camera tilts, the viewing sphere rotates relative to the upright reference frame, resulting in a corresponding geometric transformation in the ERP image, as illustrated in Fig. 1(b). Therefore, it is feasible to infer the camera's orientation (pitch and roll) from the ERP image. Traditional geometry-based methods [17–20] detect lines and estimate vanishing points under assumptions like the Manhattan world [21] or Atlanta world [22], to recover the inclination angles. However, these assumptions often fail in feature-sparse scenes (e.g., deserts) where line features are scarce. Moreover, line detection is noise-sensitive, significantly reducing estimation accuracy. In contrast, deep learning methods learn mappings from panoramic images to inclination angles using large labeled datasets without relying on explicit geometric assumptions, thereby improving adaptability to diverse scenes.

**Deep learning-based methods**: Shan et al. [23] transformed the panoramic image upright rectification task into an inclination angle classification task, and realized a step-by-step method for estimating the inclination angles of a panoramic camera from coarse to fine precision by using AlexNet [24]. Moreover, they assessed the performance of three common 2D projection images of panoramic images in the upright correction task. Jung et al. [25] presented a regression approach based on DenseNet121 for obtaining the corresponding camera inclination angles from the input non-upright panoramic image. Davidson et al. [26] integrated geometric information and vanishing points from traditional methods into image upright correction by combining image segmentation convolutional neural networks, aiming to estimate the inclination angles of the camera. Shan et al. [27] introduced a multi-scale shallow feature attention mechanism into a convolutional neural network (CNN) for estimating the camera inclination angles, thus guiding the network to pay greater attention to geometric information in the image during the learning process. Jung et al. [28] presented a method for estimating the inclination angles of a panoramic camera using a graph convolutional network (GCN). Jeon et al. [29] estimated the inclination angles of a panoramic camera by leveraging a sequence of undistorted perspective projection image patches with limited views sampled from the panoramic image in combination with a CNN. Chen et al. [14] proposed LUTgan, an end-to-end approach for generating upright panoramic images, whose training involves sequential optimization of the inclination angle estimation, 2D lookup table (LUT) generation, and upright image generation modules. Liu et al. [15] proposed PixelDSP, a U-Net-like encoder–decoder network that predicts a 2D coordinate-offset matrix for resampling non-upright ERP images into upright panoramas. More recently, the 3dCoMap framework [16] reformulated upright correction as spherical-coordinate regression, providing a three-dimensional spherical representation instead of directly predicting 2D pixel offsets. However, it still relies on features learned within a single projection domain. Such single-projection formulations may not fully exploit the complementary characteristics of different panoramic representations, motivating projection-aware representation learning across ERP and cubemap domains.

Beyond specific architectures, these learning-based techniques may also be grouped by their underlying correction strategy. **Angle-compensation methods** estimate the camera inclination and subsequently rectify the panorama through geometric transformation. Because the estimated orientation is independent of the output image resolution, these methods can be flexibly applied to different panoramic formats and resolutions without retraining the network. However, an additional geometric rectification step is required to obtain the final upright image, which may introduce extra processing overhead. In contrast, **end-to-end generation methods** directly produce upright panoramas, making them attractive for applications requiring direct image output, including immersive visual computing. Their main drawback is that errors in the predicted pixel- or spherical-coordinate mappings can introduce additional geometric distortions. Moreover, methods that directly predict resolution-dependent coordinate mappings may be tied to the training resolution, requiring adaptation or retraining for different output resolutions.

Most existing learning-based approaches rely primarily on CNNs, which are effective at capturing local spatial structures but may have limited ability to model long-range dependencies over the full panoramic field of view. In contrast, a vision transformer (ViT) is capable of learning global information through self-attention mechanisms, which is particularly crucial for panoramic images with their wide field of view. Recent advances in Transformer–CNN fusion networks [30-32] show that combining global reasoning with local structural cues can improve visual representation. Efficient multi-scale transformer architectures, such as EAPT [33], further demonstrate the value of hierarchical attention and multi-scale feature interaction for image processing. In parallel, multi-scale CNN architectures such as MFFNet [34] demonstrate that effective cross-scale feature interaction enhances detail preservation, motivating hybrid architectures that integrate complementary local and global representations.

Inspired by these findings, we propose a projection-aware dual-projection framework for 360-degree image rectification (see Fig. 2), which integrates inclination estimation and upright panorama generation within a unified architecture. Instead of relying on a single projection domain, our method exploits complementary information from ERP and cubemap representations through cross-projection feature transformation and multi-level feature fusion. The ERP-based CNN branch captures local spatial structures, whereas the cubemap-based ViT branch models global contextual information under a representation with reduced latitude-dependent distortion. Their multi-level representations are integrated through cross-projection feature transformation and adaptive fusion. The deepest ERP feature is used for inclination-angle estimation, whereas the fused features are decoded to predict a 3D spherical LUT for upright panorama generation. The network further incorporates a high-frequency enhancement module, ERP-aware padding, and channel attention to strengthen structural cues, preserve panoramic continuity, and refine fused representations. This multi-task design enables inclination estimation and panorama generation to provide complementary supervision within a unified rectification framework.

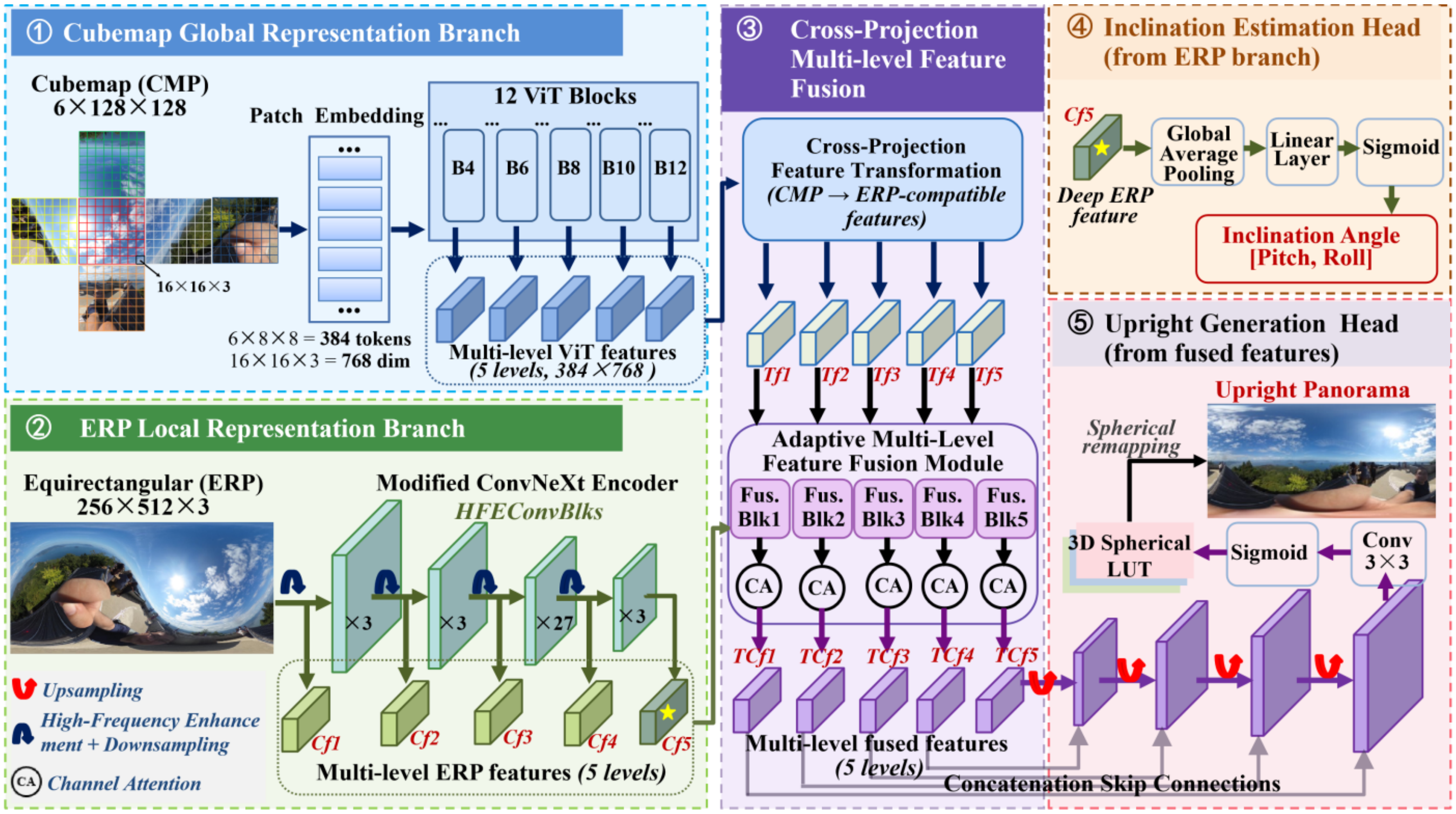

**Fig. 2: Overview of the proposed projection-aware dual-projection framework for 360-degree image rectification.** The cubemap and ERP branches extract complementary global and local representations, which are integrated through five-level cross-projection fusion. The deepest ERP feature is used for inclination angle estimation, while the fused features are decoded to predict a 3D spherical LUT for upright panorama generation.

The main contributions of this work are as follows:

1) **A dual-projection representation learning framework** is proposed for projection-aware 360-degree image rectification. It exploits complementary information from ERP and cubemap domains, where ERP-based CNN modeling captures local geometric structures and cubemap-based ViT modeling provides global contextual representations. Cross-projection feature transformation and multi-level feature fusion enable effective interaction between the two projection domains without requiring strict geometric alignment.
2) **An end-to-end multi-task learning framework** is developed to jointly estimate camera inclination angles and generate upright panoramas. The two tasks provide complementary supervision: inclination-angle supervision encourages the shared representation to capture orientation-sensitive geometric cues, while upright panorama generation provides complementary reconstruction supervision to regularize representation learning. Experiments further demonstrate that joint learning improves both tasks compared with their single-task counterparts.
3) **Panoramic-specific designs** are incorporated to enhance orientation-sensitive representation and spherical-domain modeling. These include the High-Frequency Enhancement Convolutional Block (*HFEConvBlk*) for emphasizing structural cues, ERP-aware padding for handling panoramic boundary continuity, and a 3D spherical LUT representation for spherical-domain panorama generation.
4) **Extensive experiments on two large-scale panoramic datasets, M3D [35] and SUN360 [36],** together with additional real-world evaluation on panoramas captured using a RICOH THETA camera, demonstrate the effectiveness of the proposed framework in both inclination estimation and upright panorama generation.

Ablation studies further validate the contributions of the dual-projection representation, cross-projection feature transformation and fusion, collaborative multi-task learning, and panoramic-specific designs.

## II. Motivation and Principal Analysis

While methods based on conventional geometry for estimating inclination angles and those utilizing deep learning for end-to-end image generation each have distinct advantages, they also suffer from inherent limitations. The objective of this work is to develop a projection-aware multi-task framework that simultaneously estimates camera inclination angles and generates upright panoramic images in an end-to-end manner, leveraging captured panoramic images. This approach facilitates the complementarity of advantages inherent in both tasks. Given the high degree of similarity between the objectives of these two tasks, we argue that collaborative training is feasible and has the potential to benefit both tasks.

**Inclination angle estimation** depends on the spatial structural information, such as geometric lines in the image. The equirectangular image maintains the integrity of the scene's spatial structure in the panoramic image. Therefore, it is more appropriate to implement this task by combining the equirectangular image with a CNN. The objective of the task is to infer the pitch and roll angles of the camera coordinate system *X'Y'Z'* during non-upright imaging relative to the upright imaging coordinate system *XYZ* by using the input non-upright equirectangular image, as shown in Fig. 1 (b). This is accomplished by training a network to learn the mapping relationship ***F*** between the captured non-upright equirectangular image *Img_nup* and the inclination angle *[pitch, roll]* of the camera, which is mathematically expressed as follows:

$$[pitch, roll] = F(Img_nup) \qquad (1)$$

**Upright panorama generation**'s objective is to infer the coordinate mapping that describes how each pixel in a non-upright image should be relocated to achieve an upright view. Relying solely on CNN-based local features may be insufficient for capturing this complex transformation. To achieve this, we design a dual-branch network that integrates CNN-based local structures with ViT-based global dependencies to predict a 3D spherical LUT for resampling. Unlike single-projection formulations such as PixelDSP [15] and 3dCoMap [16], the proposed framework exploits complementary representations from ERP and cubemap domains. Cross-projection feature transformation and multi-level feature fusion enable the cubemap-derived global context to interact with ERP-based local geometric features without requiring strict geometric alignment. The resulting fused representation is used to infer a 3D spherical look-up table (LUT), which provides spherical coordinates for resampling and upright panorama generation. Moreover, by adopting cubemap projection rather than equirectangular projection for the ViT branch, we reduce latitude-dependent distortions and obtain more uniform patches, which facilitates ViT-based global feature extraction.

**Cross-Projection Feature Transformation Analysis**:

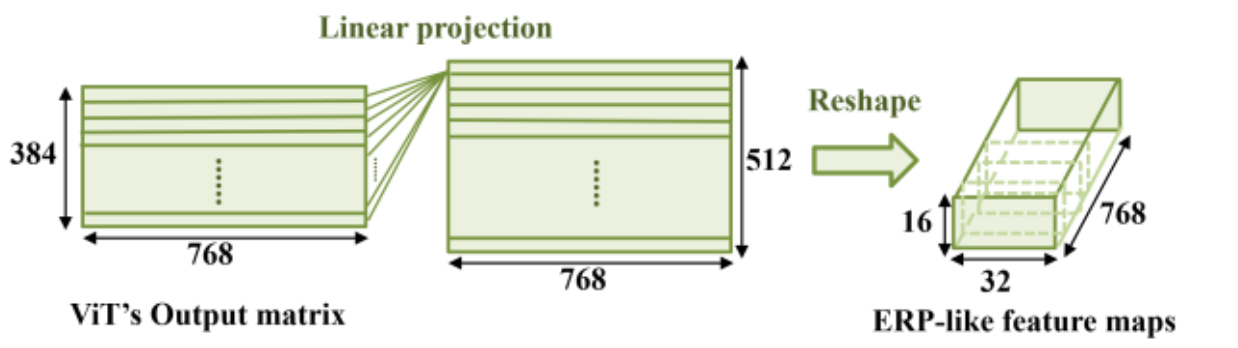

**(a)** Implicit transformation: Feature transformation via a learnable linear projection

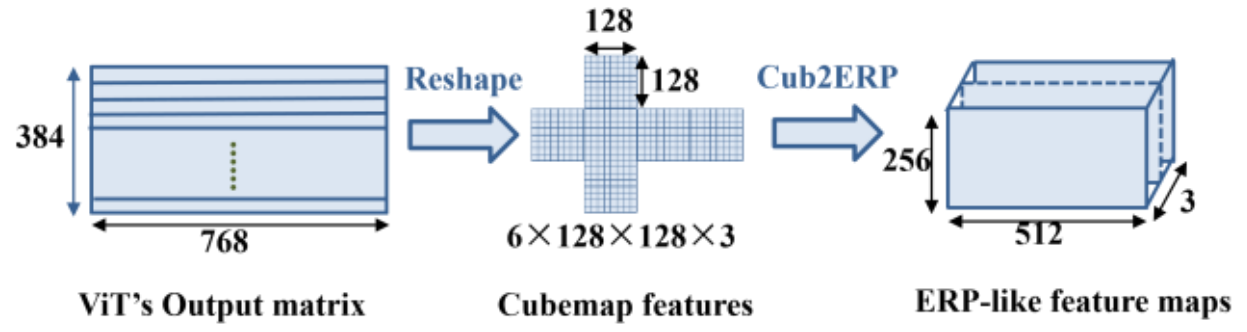

**(b)** Explicit projection: Geometric transformation via Cub2ERP

**Fig.3:** Cross-projection feature transformation strategies for feature fusion.

Since the ViT and CNN branches operate on cubemap and ERP projections respectively, each with distinct spatial distortions and discontinuities, enabling effective cross-projection feature interaction is important for cross-branch feature fusion. Directly merging features from these domains may cause semantic or spatial mismatches. To address this, we explore and compare two cross-projection transformation strategies:

1) **Implicit Data-Driven Feature Transformation:** The ViT branch encodes global contextual features into a sequence of 384 tokens, each with 768 dimensions (384=6 faces×8×8 patches; 768=16×16×3). The resulting token sequence is projected to a 512×768 matrix via a learnable linear layer and reshaped into a 16×32×768 ERP-like feature map. This allows the network to learn feature-level interaction across projections without enforcing explicit geometric alignment (Fig. 3 (a)).
   **Advantage:** Flexible and end-to-end trainable; no geometric assumptions required; lightweight (only requires a single 384 × 512 projection matrix). **Limitation:** The transformation quality depends on the network's capacity to learn effective cross-projection feature relationships from data.
2) **Explicit Geometric Projection:** The token sequence is first reshaped back into cubemap format (6 × 128 × 128 × 3), then reprojected to a 256 × 512 × 3 ERP representation via a predefined geometric projection (Cub2ERP) operation (Fig. 3 (b)).
   **Advantage:** Establishes deterministic spatial correspondence according to the known projection geometry.
   **Limitation:** Involves higher computational cost due to pixel-wise reprojection across six faces.

In both cases, ERP-like features are resized to match CNN feature map resolutions before multi-level feature fusion. Although implicit transformation lacks explicit geometric constraints, it can learn useful feature interactions across projections by leveraging statistical patterns in the data, without explicitly modeling geometric transformations. As demonstrated in our experiments, the data-driven strategy can learn compatible cross-projection representations and achieves performance comparable to the explicit projection strategy, although the two approaches rely on different underlying principles.

Based on this analysis, we adopt a dual-projection architecture: a CNN branch processes ERP inputs to estimate inclination angles and extract local features, while a ViT branch processes cubemap inputs for global feature extraction. These features are fused at multiple levels to generate a 3D spherical LUT. The final upright image is obtained by resampling the non-upright input using this LUT. Specifically, to enhance inclination estimation, we introduce a high-frequency feature enhancement module into the CNN branch to guide the network to focus more on high-frequency structural cues, such as lines, edges, and contours in the image, which are strongly related to the inclination angle estimation; For feature fusion, we adopt an adaptive fusion method to enable the network to automatically learn the relative contributions of global and local features, avoiding manually predefined fusion weights. Additionally, a channel attention module is introduced to further emphasize the features that are effective for the task and filter out irrelevant or interfering information.

## III. Inclination Estimation and Upright Panorama Generation

### *A. Overall Network Architecture*

The proposed network consists of five main components: the cubemap-based global representation branch, the ERP-based local representation branch, the cross-projection multi-level feature fusion module, the inclination estimation head, and the upright panorama generation head (Fig. 2).

### *B. Cubemap Global Representation Branch*

This module takes non-upright panoramic images in cubemap projection as input, with each face sized as 128×128. Each face is divided into 64 patches of size 16×16×3, yielding 384 patches in total. Subsequently, each patch is flattened into a one-dimensional tensor of size 1×768. Therefore, the entire panoramic image is represented by 384 one-dimensional tensors of length 768. Then, patch and positional embeddings are added, resulting in a 384×768 matrix that records the content and position information of the 384 patches.

This matrix is used as the input to the ViT encoder blocks. The number of ViT encoder blocks used in this work is 12. Each encoder contains 16 attention heads. The dimension of the hidden layer of the feedforward network is set to four times the token dimension, i.e., 3072.

The output of each ViT encoder is a matrix of size 384×768, which is consistent with the size of the input matrix. The output matrices represent global contextual representations at different encoder depths. The shallow encoders acquire fundamental visual elements, such as edges, textures, and colors, while the deeper encoders capture complex features like shapes, structures, and layouts. Deeper representations further encode long-range dependencies and contextual information. In this work, the outputs of five encoder layers with IDs [4, 6, 8, 10, 12] are selected and fused with the local features extracted by the local feature extraction module at five corresponding fusion levels to jointly predict the 3D spherical LUT used to resample the non-upright panorama into an upright panorama.

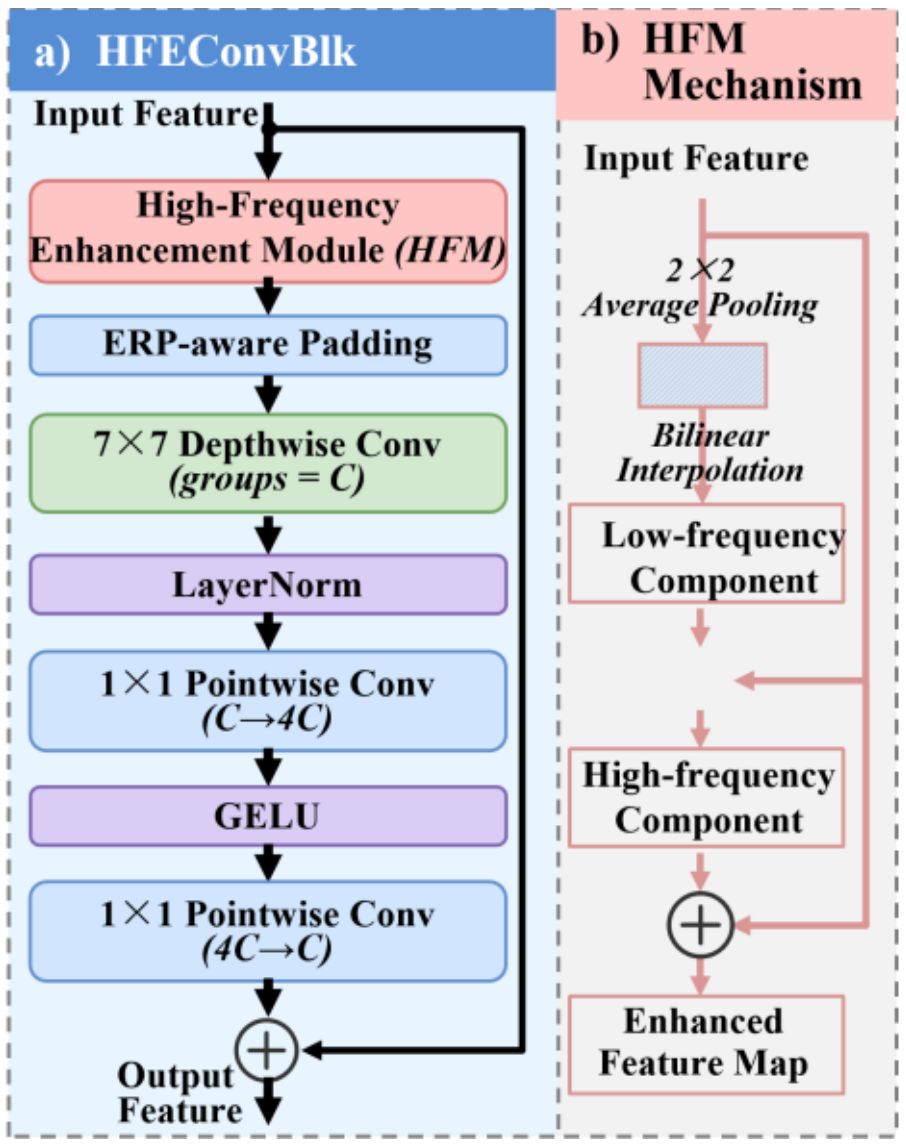

**Fig. 4. Structure of the proposed HFEConvBlk.** (a) Modified ConvNeXt block with high-frequency enhancement and ERP-aware padding. (b) Parameter-free high-frequency enhancement mechanism.

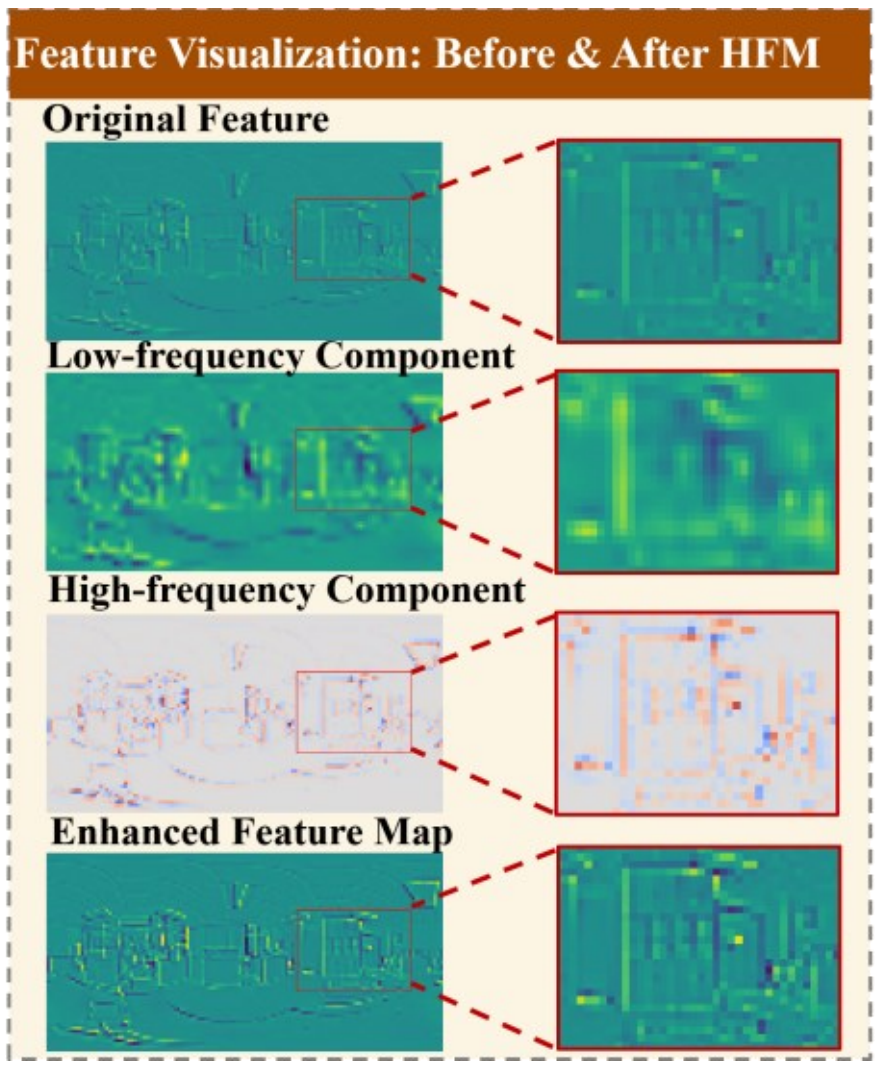

**Fig. 5. Visualization of feature responses before and after HFM.** The original, low-frequency, high-frequency, and enhanced features are shown using the same feature channel, with the same region enlarged for comparison.

### *C. ERP Local Representation Branch*

This module takes ERP-format images of size 256×512 as input, which are processed by a ConvNeXt-based backbone [37] whose basic blocks are modified by incorporating the proposed high-frequency enhancement and ERP-aware padding strategy, forming *HFEConvBlks* (Fig. 4(a)). *HFEConvBlk* introduces two key improvements: (1) a high-frequency enhancement module (HFM; Fig. 4(b)) inspired by ref. [38] that emphasizes high-frequency structural cues, such as lines, edges, and contours, which are informative for inclination estimation; and (2) an ERP-aware padding strategy designed according to the periodic horizontal structure of ERP images. Specifically, circular padding is applied at the left and right boundaries where appropriate to preserve wrap-around continuity, while conventional padding is retained along the vertical direction and in feature stages where circular padding is not adopted. As shown in Fig. 5, the HFM enhances structural contours and edge responses in the extracted feature maps, enabling the network to better exploit structural cues during training. Such edge- and contour-aware feature enhancement is consistent with recent variational energy-driven formulations for unsupervised edge detection [39], which highlight the importance of preserving high-frequency structural cues in challenging visual environments. Related structure-aware attention fusion methods have also demonstrated the effectiveness of emphasizing structural information for visual enhancement under low-quality or degraded input conditions [40].

**Table 1**
SPATIAL RESOLUTION MATCHING BETWEEN VIT AND CNN FEATURES AT FIVE FUSION LEVELS

| Fusion level | CNN Branch Resolution | ViT Branch (Implicit Transformation) (Resolution: 768×16×32) | ViT Branch (Explicit Projection) (Resolution: 3×256×512) |
|---|---|---|---|
| **Level 1** | [128×64×128] | Conv1×1: [512×16×32]<br>PixelShuffle(2): [128×32×64]<br>Conv1×1: [512×32×64]<br>PixelShuffle(2): [128×64×128] | PixelUnShuffle(2): [12×128×256]<br>Conv1×1: [32×128×256]<br>PixelUnShuffle(2): [128×64×128] |
| **Level 2** | [128×64×128] | Same as above | Same as above |
| **Level 3** | [256×32×64] | Conv1×1: [1024×16×32]<br>PixelShuffle(2): [256×32×64]<br>Conv1×1: [256×32×64] | PixelUnShuffle(4): [48×64×128]<br>Conv1×1: [64×64×128]<br>PixelUnShuffle(2): [256×32×64] |
| **Level 4** | [512×16×32] | Conv1×1: [512×16×32] | PixelUnShuffle(8): [192×32×64]<br>Conv1×1: [128×32×64]<br>PixelUnShuffle(2): [512×16×32] |
| **Level 5** | [1024×8×16] | Conv2×2 (stride2): [1024×8×16] | PixelUnShuffle(8): [192×32×64]<br>Conv1×1: [64×32×64]<br>PixelUnShuffle(4): [1024×8×16] |

As discussed in Sec. II, the extracted local spatial structure information is utilized to estimate the camera inclination angles. Additionally, in upright panoramic image generation, this local spatial structure information is combined with global contextual features to predict the 3D spherical LUT used for panorama resampling. As illustrated in the local feature extraction branch in Fig. 2, the output of the final ConvNeXt stage, denoted as *Cf5*, is used to estimate the camera inclination angles. These feature maps undergo global average pooling, linear projection, and sigmoid activation, producing normalized [pitch, roll] inclination angles. These angles are transformed into angular values as follows:

$$[P_{ang}, R_{ang}] = [pitch, roll] \times 180 - 90 \qquad (2)$$

The reasons for using only the local spatial structure features for predicting the inclination angles are as follows: Inclination angle estimation relies heavily on the geometric information within the image. Convolution operations excel at extracting pixel-level features while preserving the overall spatial structure of the scene during feature extraction. The ViT branch, in contrast, is primarily introduced to capture complementary global contextual information from the cubemap representation and is therefore used in the panorama generation pathway rather than directly in the inclination estimation head. It is worth noting that although the global contextual features extracted by ViT do not directly contribute to inclination angle estimation, the joint generation task indirectly affects the optimization of the convolutional branch through shared multi-task training.

### *D. Adaptive Multi-Level Feature Fusion Module*

To effectively integrate the multi-level global features extracted by the ViT encoder (from transformer blocks 4, 6, 8, 10, and 12) with the hierarchical local features produced by the CNN encoder (i.e., the output after the initial downsampling and the outputs of the four ConvNeXt stages), we design an adaptive multi-scale fusion module composed of the following four processing steps:

**1) Cross-Projection Feature Transformation:** To enable integration with ERP-based CNN features, the ViT token sequence is transformed into an ERP-compatible representation using either the implicit data-driven transformation or explicit geometric projection described in Sec. II and Fig. 3. The resulting ERP-like features are then used for subsequent multi-level fusion.

**2) Spatial Resolution Matching:** To ensure compatible spatial resolutions between features from the ViT and CNN branches, the ERP-like features are resized to match the spatial dimensions of the corresponding CNN feature maps. We utilize 1×1 convolution layers combined with PixelShuffle upsampling for resolution enhancement, and PixelUnshuffle or 2×2 convolution (stride 2) for downsampling. The specific resolution matching configurations across the five fusion levels are summarized in Table1.

**3) Adaptive Global–Local Feature Fusion:** Following the principle of fusing features at corresponding representation levels (Fig. 2), we introduce a residual-enhanced adaptive fusion module (Fig. 6) inspired by ref. [41]. At each fusion level, global features from the ViT branch (denoted as *Tf*) and local features from the CNN branch (denoted as *Cf*) are integrated through a spatial attention map generated by convolutional layers. This attention map guides the weighted summation of *Tf* and *Cf*, while residual connections help preserve the original feature information and stabilize the fused representations. The fusion process can be formulated as follows; the initial joint representation is first obtained by element-wise aggregation:

$$TCf_{sum} = Tf + Cf \tag{3}$$

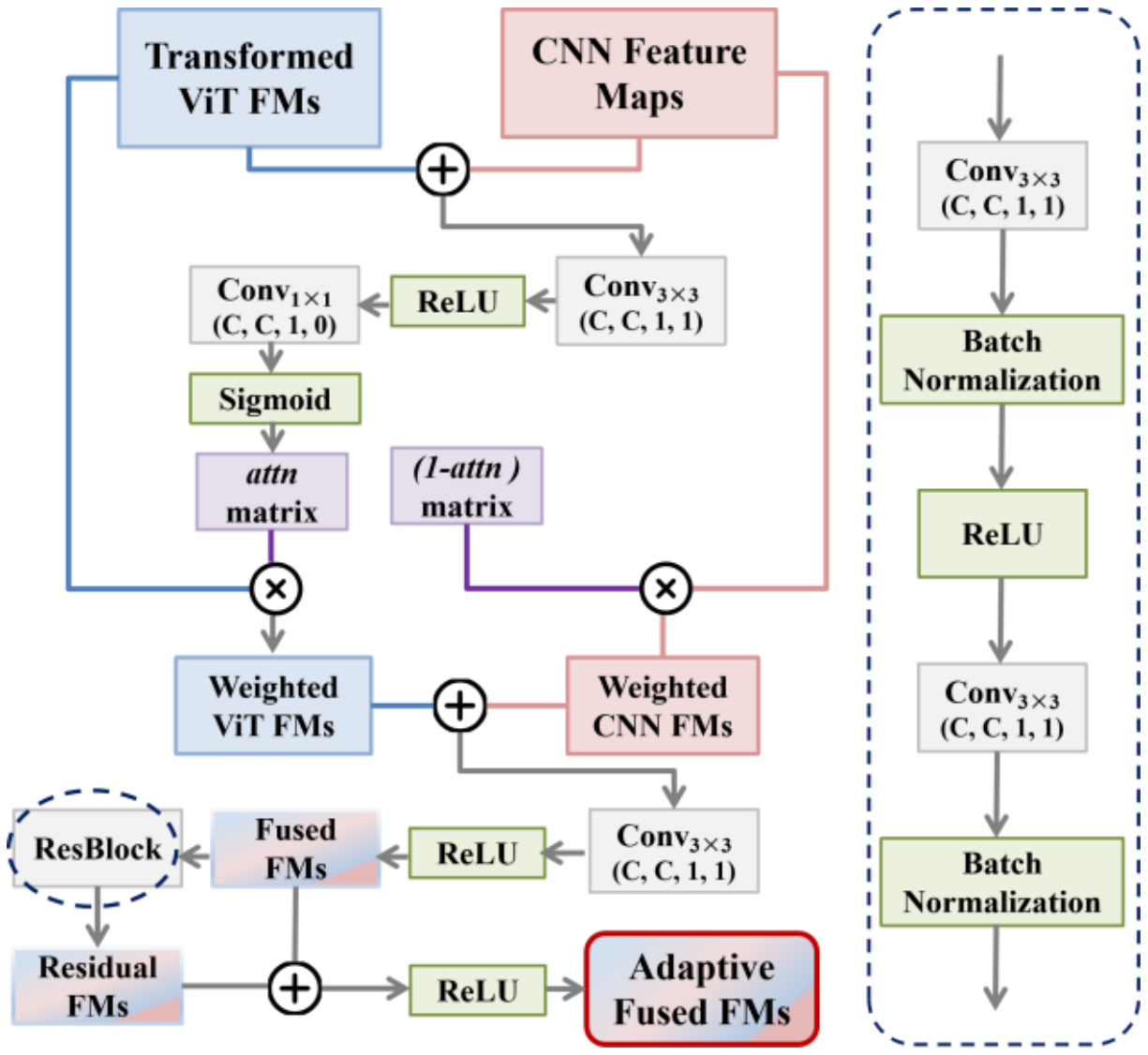

**Fig. 6:** Architecture of the Adaptive Global–Local Feature Fusion Module. (*FMs: feature maps; (C, C, 1, 1) denotes convolution parameters—input channels, output channels, kernel size, and padding. Circular padding is applied to the left and right boundaries of the feature maps.*)

It is then processed by a 3×3 convolution, *ReLU* activation, and a 1×1 convolution, followed by a sigmoid function to generate the spatial attention map:

$$TCf_{enc} = Conv_1(relu(Conv_3(TCf_{sum}))) \tag{4}$$

$$attn = Sigmoid(TCf_{enc}) \tag{5}$$

The learned attention map adaptively weights the global and local representations:

$$Tf_w = Tf \odot attn \tag{6}$$

$$Cf_w = Cf \odot (1 - attn) \tag{7}$$

$$TCf_{fused} = Tf_w + Cf_w \tag{8}$$

Finally, the fused representation is refined through a convolutional residual block:

$$TCf_{fused} = relu(Conv_3(TCf_{fused})) \tag{9}$$

$$TCf_{res} = BN(Conv_3(relu(BN(Conv_3(TCf_{fused}))))) \tag{10}$$

$$TCf_{final} = relu(TCf_{fused} + TCf_{res}) \tag{11}$$

where *BN(·)* denotes batch normalization.

This design allows the network to learn spatially adaptive weightings between global and local features while improving the discriminative capability of the fused representations through residual refinement.

**4) Channel-wise Refinement:** To suppress less informative responses and emphasize task-relevant channels after fusion, we apply a lightweight channel attention (CA) mechanism. Global average pooling and two 1×1 convolutions are used to generate channel-wise attention weights, which are subsequently applied to refine the fused features.

This fusion module enables effective integration of global contextual information and local structural cues across multiple representation levels, providing the fused representations used by the decoder for 3D spherical LUT prediction and upright panorama generation. The inclination estimation head, in contrast, directly operates on the deepest ERP feature *Cf5*, while the two tasks remain jointly coupled through the shared CNN representation during multi-task training.

### *E. Upright Panorama Generation Head*

As depicted in the "Upright Panorama Generation Head" in Fig. 2, the decoder progressively upsamples the deepest fused representation *TCf5* and incorporates the corresponding fused features *TCf4* – *TCf1* through concatenation-based skip connections. Each decoding stage applies convolutional transformation and ELU activation, followed by nearest-neighbor upsampling where required. At the final feature resolution (64×128), subpixel convolution (PixelShuffle) is used to upscale the feature maps to 256×512. A final 3×3 convolution followed by a sigmoid-based coordinate transformation produces a 256×512×3 3D spherical LUT, which specifies the predicted spherical sampling coordinates. The upright panorama is then obtained by resampling the non-upright input according to the predicted LUT. Detailed layer-by-layer decoding operations are provided in Algorithm S1 of the Supplementary Material.

### *F. Loss function*

The loss function used in this work is composed of three main components: inclination angle loss, spherical LUT loss, and perceptual loss for upright panorama generation.

The **inclination angle loss** includes two terms: (1) a smooth L1 loss between the predicted inclination angles and the ground truth values, and (2) an L1 loss between the upright image corrected by the predicted angles and the ground truth upright image. This is formulated as:

$$L_{Ang} = L_{SmoothL1}(A_{pred}, A_{gt}) + L_1(I_{Ang}, I_{gt}) \tag{12}$$

where $A_{pred}$ and $A_{gt}$ denote the predicted and ground-truth pitch–roll angle vectors, respectively, $I_{Ang}$ denotes the upright panorama obtained using the predicted inclination angles, and $I_{gt}$ denotes the ground-truth upright panorama.

The **LUT-based geometric loss** consists of three components: (1) a mean squared error (MSE) loss between the predicted 3D spherical LUT and its ground truth; (2) an L1 reconstruction loss between the panorama resampled using the predicted LUT and the ground-truth upright panorama; and (3) a unit-sphere constraint that encourages each predicted 3D sampling coordinate to lie on the unit sphere. It is expressed as:

$$L_{LUT} = L_{MSE}(L_{pred}, L_{gt}) + L_1(I_{LUT}, I_{gt}) + L_{sphere} \tag{13}$$

where $L_{pred}$ and $L_{gt}$ denote the predicted and ground-truth 3D spherical LUTs, respectively, and $I_{LUT}$ denotes the upright panorama obtained by resampling the non-upright input using the predicted LUT. The unit-sphere constraint is defined as

$$L_{sphere} = L_1(\|L_{pred}\|_2^2, 1) \tag{14}$$

where the squared Euclidean norm is computed along the three coordinate channels at each spatial location, and **1** is an all-ones matrix with the same spatial dimensions as the predicted LUT. This constraint encourages each predicted coordinate $(x, y, z)$ to satisfy $x^2 + y^2 + z^2 = 1$.

The **perceptual loss** is introduced to improve the perceptual similarity of the generated upright panoramas. We adopt Learned Perceptual Image Patch Similarity (LPIPS), which measures perceptual differences using deep feature representations:

$$L_{perceptual} = L_{LPIPS}(I_{LUT}, I_{gt}) \quad (15)$$

Finally, the **overall training objective** is defined as

$$L_{total} = \alpha \cdot L_{Ang} + \beta \cdot L_{LUT} + \gamma \cdot L_{perceptual} \quad (16)$$

where $\alpha, \beta, \gamma$ are balancing weights to account for the different magnitudes of the losses.

Each loss component provides complementary supervision for different aspects of upright panorama correction. Specifically, the inclination angle loss constrains camera orientation estimation, while the spherical LUT loss supervises spherical-coordinate prediction and geometric reconstruction between non-upright and upright panoramas. Within the LUT loss, the reconstruction term promotes image-level geometric consistency, whereas the unit-sphere constraint regularizes the validity of the predicted spherical sampling coordinates. The perceptual loss further complements these geometric objectives by improving the perceptual similarity of the generated upright panorama.

## IV. Experiments and Discussions

### A. Dataset and Experiments Settings

To ensure fair evaluation, we test our method on two widely used public panoramic datasets, **M3D** and **SUN360**, which have been commonly adopted in previous upright panorama rectification studies. M3D contains over 50,000 indoor upright equirectangular images, while SUN360 contains approximately 70,000 indoor and outdoor panoramas with greater scene diversity. Both datasets contain panoramic images from diverse real-world scenes, covering substantial variations in scene content, illumination, structural complexity, and visual quality.

Following common practice in previous works, we use a resolution of 256×512 for all equirectangular images to maintain consistency and ensure fair comparison. Each original upright panorama is used to generate one non-upright sample with a single randomly sampled pitch–roll pair. The resulting dataset is then randomly divided into training, validation, and testing subsets at a ratio of 70%, 15%, and 15%, respectively. The training, validation, and test subsets are disjoint at the source-panorama level, with no original panorama appearing in more than one subset. The data preparation procedure consists of the following five steps:

1) **Upright equirectangular images:** All original upright panoramic images are downsampled to 256×512 resolution and used as ground-truth upright images.
2) **Ground-truth inclination angles:** For each image, random pitch and roll angles are generated within the range [–90°, 90°] to simulate camera inclination. These angles serve as the ground truth for inclination angle estimation.
3) **Non-upright equirectangular images:** Each upright equirectangular image is first projected onto a unit sphere (as illustrated in Fig. 1 (b)), then rotated using the ground-truth pitch and roll angles to simulate a tilted camera. The rotated sphere is subsequently reprojected to the equirectangular format to produce the non-upright input image, which is fed into the ERP local representation branch (see Fig. 2). The rotation is defined as

$$\begin{bmatrix} x' \\ y' \\ z' \end{bmatrix} = \begin{bmatrix} \cos(p) & \sin(p)\sin(r) & \sin(p)\cos(r) \\ 0 & \cos(r) & -\sin(r) \\ -\sin(p) & \cos(p)\sin(r) & \cos(p)\cos(r) \end{bmatrix} \begin{bmatrix} x \\ y \\ z \end{bmatrix} \quad (17)$$

where, $(x, y, z)$ denotes the unit sphere coordinates of the pixels in the upright equirectangular image. These coordinates are left-multiplied by the rotation matrix derived from the inclination angles $p$ (pitch) and $r$ (roll), resulting in the unit-sphere coordinates $(x', y', z')$ of the corresponding pixels in the non-upright equirectangular image. This transformation simulates the effect of camera inclination on the spherical projection of the panoramic image.
4) **Non-upright cubemap projection:** Each non-upright image is converted into a cubemap format consisting of six 128×128 faces, which is used as input to the cubemap global representation branch (see Fig. 2).
5) **Ground-truth 3D spherical LUT:** For each non-upright image, we compute a 256×512×3 ground-truth spherical LUT, which specifies the spherical sampling coordinates used to reconstruct the upright panorama. This is derived using the inverse transformation of Eq. (17), with the north pole coordinate (0, 0, 1) on the unit sphere used as the upright reference.

For the overall loss in Eq. (16), we set $\alpha = 1, \beta = 10, \gamma = 10$ based on preliminary experiments to balance the numerical scales of the different loss components. The Adam optimizer is employed with an initial learning rate of 0.0001. A batch size of 6 is used, and the random seed is set to 100. The proposed network is trained for 100 epochs using an NVIDIA RTX 3090 GPU.

To evaluate the accuracy of predicted inclination angles, we report the percentage of test samples whose predicted pitch and roll angles fall within angular error thresholds of [1°, 2°, 3°, 4°, 5°, 12°]. We further report mean absolute error (MAE), root mean square error (RMSE), median angular error, 95th-percentile angular error (P95), Geo-MAE, and Geo-RMSE for a more comprehensive evaluation, where Geo-MAE and Geo-RMSE are the mean and root-mean-square SO(3) geodesic errors between predicted and ground-truth pitch–roll orientations with yaw fixed to zero.

To evaluate upright panorama generation, we adopt both perceptual and pixel-level image quality metrics. Fréchet Inception Distance (FID) and Learned Perceptual Image Patch Similarity (LPIPS) are used to assess perceptual similarity, while Normalized Root Mean Square Error (NRMSE), Normalized Mean Absolute Error (NMAE), Peak Signal-to-Noise Ratio (PSNR), and Structural Similarity Index Measure (SSIM) are used to evaluate reconstruction fidelity with respect to the ground-truth upright panoramas.

### B. Comparison with Prior Approaches

To better demonstrate the effectiveness of our method, we evaluate two variants of the proposed network based on different cross-projection feature transformation strategies:

**Ours-Exp** denotes the version using explicit geometric projection, where cubemap features are reprojected onto ERP via predefined transformations. **Ours-Imp** adopts an implicit data-driven transformation, where cross-projection feature interaction is learned through a learnable linear transformation without explicit geometric constraints. Both variants are consistently evaluated across all benchmarks and compared with representative existing methods.

This setting allows us to benchmark against prior work while also examining the performance differences between explicit geometric projection and implicit data-driven transformation.

**Table 2.**
A COMPARISON OF INCLINATION ANGLE PREDICTION ACCURACY (@SUN360, WITHIN ±60° RANGE).

| Methods | Accuracy within Angular-Error Thresholds (%). | | | | | |
|---|---|---|---|---|---|---|
| | <1° | <2° | <3° | <4° | <5° | <12° |
| **Cos2Fine** | 30.1 | 51.7 | 65.9 | 74.0 | 79.1 | 91.0 |
| **De360Up** | 7.1 | 24.5 | 43.9 | 60.7 | 74.2 | 97.9 |
| **vp-gscnn** | 19.7 | 53.6 | 75.5 | 87.2 | 92.6 | 98.4 |
| **LUTgan** | 29.9 | 65.3 | 80.3 | 86.3 | 89.2 | 95.2 |
| **MSAtten** | 31.2 | 72.9 | 89.8 | 95.5 | 97.5 | 99.5 |
| **3dCoMap** | 53.3 | 83.0 | 90.4 | 93.5 | 94.9 | 97.7 |
| **Ours-Exp** | 61.2 | 93.6 | 99.0 | 99.6 | 99.8 | **100** |
| **Ours-Imp** | **65.9** | **95.5** | **99.0** | **99.7** | **99.9** | **100** |

**Table 3.**
A COMPARISON OF PREDICTION ACCURACY FOR DIFFERENT INCLINATION ANGLE RANGES (@ SUN360).

| Methods | | Accuracy within Angular-Error Thresholds (%). | | | | | |
|---|---|---|---|---|---|---|---|
| | | <1° | <2° | <3° | <4° | <5° | <12° |
| **MsAtten** | ±30° | 35.3 | 77.0 | 91.5 | 95.8 | 97.6 | 99.3 |
| | ±60° | 31.2 | 72.9 | 89.8 | 95.5 | 97.5 | 99.5 |
| | ±90° | 26.9 | 62.1 | 77.8 | 84.6 | 88.0 | 94.6 |
| **Ours-Exp** | ±30° | 63.2 | 92.7 | 98.8 | 99.6 | 99.7 | 100 |
| | ±60° | 61.2 | 93.6 | 99.0 | 99.6 | 99.8 | 100 |
| | ±90° | 53.8 | 82.7 | 89.7 | 92.1 | 93.5 | 97.0 |
| **Ours-Imp** | ±30° | 63.8 | 95.4 | 98.6 | 99.6 | 99.8 | 100 |
| | ±60° | 65.9 | 95.5 | 99.0 | 99.7 | 99.9 | 100 |
| | ±90° | 57.2 | 84.0 | 89.9 | 92.4 | 93.6 | 96.9 |

### 1) Inclination Angle Estimation Task

**SUN360 dataset:** We compare our method with Cos2Fine [23], De360Up [25], vp-gscnn [26], LUTgan [14], MSAtten [27], and 3dCoMap [16] which have reported inclination angle estimation performance on this dataset. Notably, Cos2Fine and vp-gscnn restrict their evaluation to within ±60°, so for fairness, we adopt the same range for all methods. As shown in Table 2, our method achieves higher accuracy than the compared methods across the reported angular error thresholds, with particularly clear improvements under stricter thresholds.

MSAtten trained the network over an inclination range of ±90°. The prediction accuracies for non-upright panoramic images within the ranges of ±30°, ±60°, and ±90° are presented in Table 3. The reported results show relatively stable performance within ±60°, followed by a noticeable decrease when the inclination range is extended to ±90°. A similar trend is observed for the proposed method. Notably, when evaluated over the full ±90° range, our method still achieves higher accuracy than MSAtten evaluated over ±60° at the strict 1° and 2° error thresholds. The box-and-whisker plots in Fig. 7 further illustrate the distribution of inclination angle prediction errors across 10° inclination intervals. The errors remain relatively low up to the 70°–80° range, but increase sharply when the inclination approaches ±90°. This increase may be partly associated with the ambiguity and sensitivity of Euler-angle representations near extreme orientations, where different pitch–roll combinations can correspond to similar upright configurations. Extreme ERP orientations may also make structural cues more difficult to interpret. As further illustrated by the challenging-case analysis in Section IV-E, relatively large pitch–roll discrepancies near ±90° do not necessarily indicate unsuccessful visual rectification.

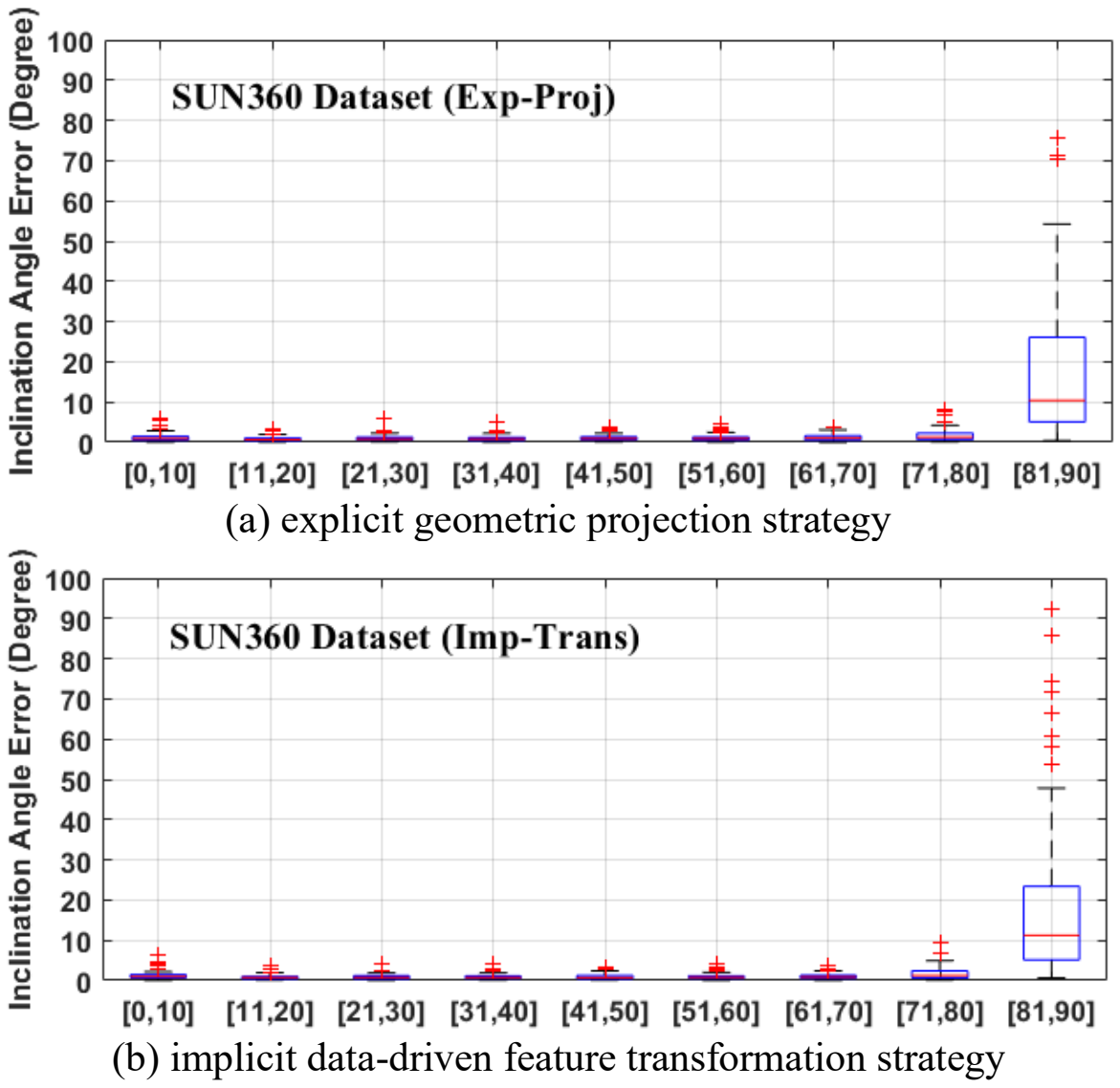

**Fig. 7.** Box plots of angle error distributions across different camera inclination ranges (in degrees) on the SUN360 dataset.

**Table 4.**
A COMPARISON OF INCLINATION ANGLE PREDICTION ACCURACY (@ M3D, WITHIN ±90° RANGE).

| Methods | Accuracy within Angular-Error Thresholds (%). | | | | | | Average error angle (°) |
|---|---|---|---|---|---|---|---|
| | <1° | <2° | <3° | <4° | <5° | <12° | |
| **PixelDSP** | - | - | - | - | - | - | 1.807 |
| **MsAtten** | 70.3 | 89.2 | 93.0 | 94.5 | 95.5 | 97.4 | 0.848 |
| **3dCoMap** | 66.2 | 90.3 | 94.8 | 96.3 | 97.1 | 98.6 | - |
| **Ours-Exp** | 82.5 | 94.5 | 96.5 | 97.2 | 97.6 | 98.6 | 0.659 |
| **Ours-Imp** | **85.2** | **95.0** | **97.0** | **97.7** | **98.0** | **98.8** | **0.614** |

**Table 5.**
ADDITIONAL ANGULAR-ERROR STATISTICS (WITHIN ±90° RANGE)

| Data | Var. | MAE↓ | RMSE↓ | Med.↓ | P95↓ | Geo-MAE↓ | Geo-RMSE↓ |
|---|---|---|---|---|---|---|---|
| **SUN** | Exp. | 2.420 | 8.384 | 0.932 | 6.656 | 2.451 | 8.802 |
| | Imp. | **2.280** | **7.814** | **0.869** | **6.430** | **2.326** | **8.459** |
| **M3D** | Exp. | 1.467 | **7.435** | 0.534 | 2.142 | 1.513 | 8.148 |
| | Imp. | **1.398** | 7.804 | **0.485** | **2.014** | **1.414** | **8.090** |

**M3D Dataset:** We compare with 3dCoMap [16], MSAtten [27] and PixelDSP [15], where PixelDSP primarily focuses on end-to-end upright image generation and does not directly

**Table 6.**
QUANTITATIVE COMPARISON OF UPRIGHT PANORAMA RECTIFICATION QUALITY (WITHIN ±90° RANGE).

| Data | Type | Method | Output Resolution | FID↓ | NRMSE↓ | NMAE↓ | PSNR ↑ | SSIM ↑ | LPIPS ↓ |
|---|---|---|---|---|---|---|---|---|---|
| SUN360 | Learned generation | LUTgan | 256×512 | 53.32 | 0.3225 | 0.2180 | - | - | - |
| | | Ours-Exp | 256×512 | <u>5.56</u> | **0.1811** | **0.1151** | **21.64** | <u>0.642</u> | **0.115** |
| | | Ours-Imp | 256×512 | **5.87** | <u>0.1859</u> | <u>0.1182</u> | <u>21.58</u> | **0.638** | <u>0.116</u> |
| | Angle-based geometric rectification | Ours (Angle-Geo) | 256×512 | <u>6.56</u> | **0.1816** | **0.1160** | **21.73** | **0.644** | **0.119** |
| | | Ours (Angle-Geo) | 512×1024 | **3.47** | <u>0.1935</u> | <u>0.1261</u> | <u>21.23</u> | <u>0.585</u> | <u>0.214</u> |
| M3D | Learned generation | PixelDSP | 256×512 | 24.49 | 0.2268 | 0.1363 | 24.70 | 0.774 | 0.109 |
| | | 3dCoMap | 256×512 | 18.40 | **0.1512** | **0.0829** | - | - | - |
| | | Ours-Exp | 256×512 | **2.98** | <u>0.1684</u> | <u>0.1101</u> | **27.25** | **0.838** | **0.054** |
| | | Ours-Imp | 256×512 | <u>3.26</u> | 0.1728 | 0.1128 | <u>26.97</u> | <u>0.832</u> | <u>0.055</u> |
| | Angle-based geometric rectification | Ours (Angle-Geo) | 256×512 | <u>2.45</u> | <u>0.1534</u> | <u>0.0924</u> | <u>27.92</u> | **0.855** | **0.051** |
| | | Ours (Angle-Geo) | 512×1024 | **1.66** | **0.1529** | **0.0901** | **28.04** | <u>0.846</u> | <u>0.080</u> |

* ↑ indicates higher is better, while ↓ indicates lower is better. Best and second-best results within each method category are shown in bold and underlined, respectively. "–" indicates that the corresponding metric was not reported or is unavailable.

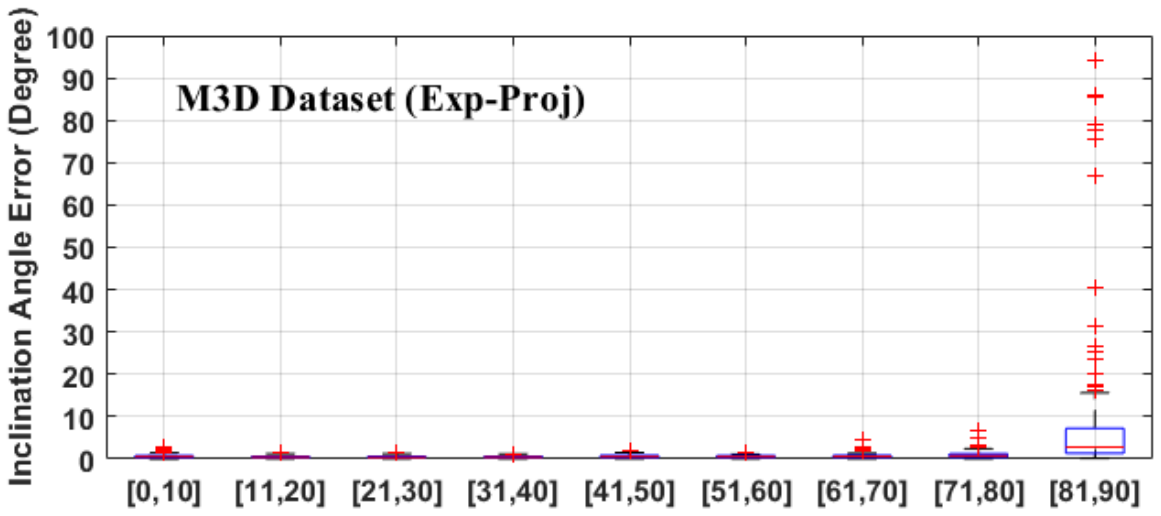

(a) explicit geometric projection strategy

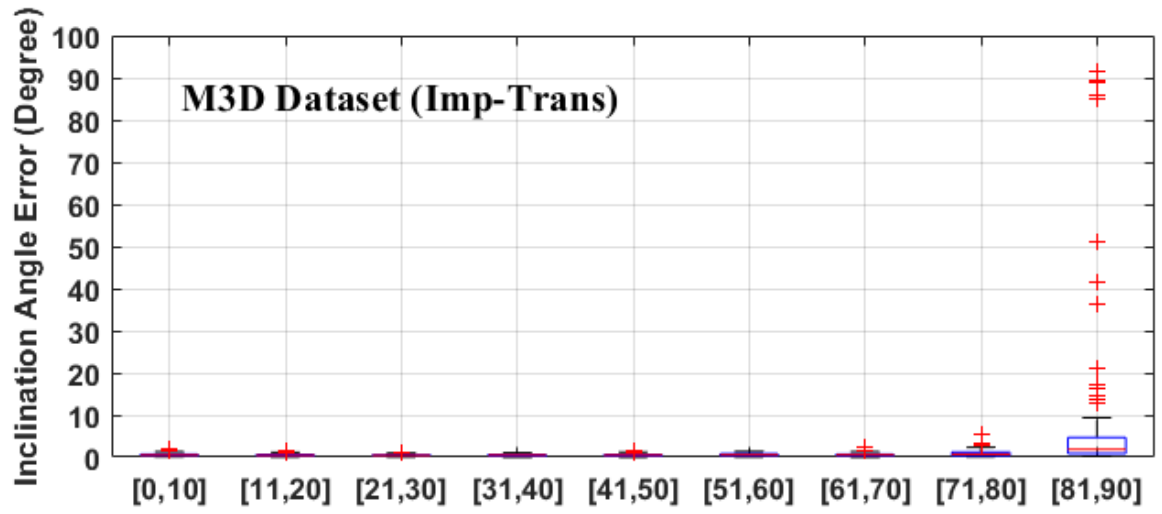

(b) implicit data-driven feature transformation strategy

**Fig. 8.** Box plots of angle error distributions across different camera inclination ranges (in degrees) on the M3D dataset.

predict inclination angles. To enable comparison, PixelDSP reports a coarse evaluation based on the average error of samples whose estimation error is below 5°, and we follow the same protocol. As shown in Table 4, the proposed method achieves higher inclination estimation accuracy than the compared methods across the reported angular error thresholds. Compared with the SUN360 results in Table 2, the proposed method achieves higher inclination estimation accuracy on M3D. A possible explanation is that M3D contains only indoor scenes, where structural cues such as lines and regular layouts are generally more abundant than in the more diverse SUN360 dataset. A similar trend is observed on M3D (Fig. 8), although the error increase near extreme inclinations is less pronounced than on SUN360.

**Additional angular-error statistics:** Table 5 reports the MAE, RMSE, median error, 95th-percentile error (P95), and SO(3) geodesic errors over the full ±90° test range. On SUN360, Ours-Imp achieves lower errors than Ours-Exp across all reported statistics, including MAE (2.280° vs. 2.420°), median error (0.869° vs. 0.932°), and P95 (6.430° vs. 6.656°). On M3D, Ours-Imp also yields lower MAE, median, P95, and geodesic errors, whereas Ours-Exp gives a slightly lower RMSE (7.435° vs. 7.804°). This difference suggests that a small number of large-error samples affect the RMSE more strongly, while the lower median and P95 values indicate better overall accuracy for the majority of samples.

### 2) Upright Panorama Generation and Rectification Quality

Among prior works, LUTgan [14], PixelDSP [15], and 3dCoMap [16] perform end-to-end upright panoramic image generation. Table 6 compares their rectification quality with our learned-generation and angle-based geometric rectification routes using FID, NRMSE, NMAE, PSNR, SSIM, and LPIPS where available. For LUTgan, we report the quantitative results available from PixelDSP, while the PixelDSP outputs and corresponding ground-truth panoramas provided for its test set are re-evaluated using the same metric implementation as our method.

Notably, 3dCoMap predicts upright panoramas through a learned 3D coordinate mapping matrix. Although it achieves slightly lower NRMSE and NMAE than our method, its inclination estimation accuracy and perceptual quality are lower (66.2% vs. 82.5%/85.2% within 1°, see Table 4; FID 18.40 vs. 2.98/3.26). This indicates that pixel-level reconstruction and perceptual/orientation quality reflect different aspects of panorama rectification. Compared with methods that directly learn a single coordinate-mapping representation, our framework jointly models inclination estimation and upright panorama generation while exploiting complementary ERP and cubemap features through cross-projection fusion, which helps explain the improved orientation accuracy and perceptual quality.

For the two learned generation variants, Ours-Exp achieves slightly better image-generation quality than Ours-Imp on both datasets. Ours-Exp generally obtains lower perceptual and reconstruction errors, while the performance gap remains relatively small, indicating that both explicit geometric projection and implicit data-driven transformation can effectively support cross-projection feature fusion.

**Table 7.**
THE ABLATION ANALYSIS OF INCLINATION ANGLE PREDICTION ACCURACY AND END-TO-END UPRIGHT IMAGE GENERATION QUALITY (@M3D).

| OriConvNeXt | ERP Pad. | HFM | CA | ViT | Accuracy within Angular-Error Thresholds (%). | | | | | | Quality of image generation | | |
|---|---|---|---|---|---|---|---|---|---|---|---|---|---|
| | | | | | <1° | <2° | <3° | <4° | <5° | <12° | FID↓ | NRMSE↓ | NMAE↓ |
| × | × | × | × | √ | 0.6 | 1.8 | 3.7 | 6.1 | 8.9 | 34.6 | 50.98 | 0.3405 | 0.2374 |
| √ | × | × | × | × | 81.5 | 94.1 | 96.2 | 97.2 | 97.6 | 98.5 | 5.17 | 0.1991 | 0.1278 |
| √ | √ | × | × | × | 82.1 | 94.2 | 96.4 | 97.1 | 97.6 | 98.5 | 5.07 | 0.1902 | 0.1229 |
| √ | √ | √ | × | × | 85.8 | 94.6 | 96.7 | 97.4 | 97.8 | 98.7 | 4.79 | 0.1842 | 0.1191 |
| √ | √ | √ | √ | × | **86.4** | **95.2** | **97.0** | **97.7** | **98.0** | **98.8** | 3.95 | 0.1809 | 0.1170 |
| √ | √ | √ | √ | √ | 85.2 | 95.0 | **97.0** | **97.7** | **98.0** | **98.8** | **3.26** | **0.1728** | **0.1128** |

**Table. 8.**
MUTUAL REINFORCEMENT BETWEEN INCLINATION ESTIMATION AND UPRIGHT IMAGE GENERATION (@M3D)

| Model Variant | Accuracy within Angular-Error Thresholds (%). | | | | | | Quality of image generation | | |
|---|---|---|---|---|---|---|---|---|---|
| | <1° | <2° | <3° | <4° | <5° | <12° | FID↓ | NRMSE↓ | NMAE↓ |
| **Only-Angle** | 81.9 | 94.3 | 96.4 | 97.3 | 97.6 | 98.7 | - | - | - |
| **Only-Image** | - | - | - | - | - | - | 3.68 | **0.1716** | **0.1119** |
| **Full Model (Imp-Trans)** | **85.2** | **95.0** | **97.0** | **97.7** | **98.0** | **98.8** | **3.26** | 0.1728 | 0.1128 |

**Table. 9.**
ABLATION STUDY ON PROJECTION DESIGN (@M3D)

| Model Variant | Accuracy within Angular-Error Thresholds (%). | | | | | | Quality of image generation | | |
|---|---|---|---|---|---|---|---|---|---|
| | <1° | <2° | <3° | <4° | <5° | <12° | FID↓ | NRMSE↓ | NMAE↓ |
| **ERP + ERP** | 83.7 | 94.7 | 96.9 | 97.6 | 98.0 | 98.7 | 3.42 | 0.1733 | 0.1129 |
| **ERP + CMP (Ours Imp-Trans)** | **85.2** | **95.0** | **97.0** | **97.7** | **98.0** | **98.8** | **3.26** | **0.1728** | **0.1128** |

In addition to the learned generation route, Table 6 reports Ours (Angle-Geo), where the predicted pitch and roll are directly applied through deterministic spherical resampling. Since this route does not involve a learned image decoder, it is analyzed separately from the end-to-end generation methods.

Angle-Geo achieves competitive rectification quality, particularly on M3D, which is consistent with the higher inclination estimation accuracy obtained on this structurally regular indoor dataset. Furthermore, the predicted inclination angles from 256×512 inputs can be directly transferred to 512×1024 panoramas without retraining, demonstrating the resolution-independent applicability of the angle-based correction strategy. The corresponding runtime analysis is presented in Sec. IV-F.

Overall, the proposed framework achieves competitive performance compared with existing approaches. Two main design differences distinguish our method: (1) unlike LUTgan [14] and PixelDSP [15], which predict 2D offsets in ERP space, our method uses a 3D spherical LUT for panorama resampling; compared with 3dCoMap [16], which directly learns a 3D coordinate mapping matrix for upright adjustment, our method additionally introduces explicit inclination estimation and exploits complementary ERP and cubemap representations through cross-projection interaction and fusion; and (2) global contextual information and local structural cues are jointly integrated to improve spherical coordinate transformation and upright panorama generation.

Considering the trade-off between the two cross-projection strategies, Ours-Exp provides slightly better panorama-generation quality, whereas Ours-Imp achieves more accurate inclination estimation under strict thresholds and lower MAE, median error, P95, and geodesic errors (Table 5). Since Ours-Imp also reduces computational overhead (Sec. IV-F), avoids explicit reprojection, and maintains comparable generation quality, it is adopted for subsequent ablation studies.

### *C. Ablation Analysis*

#### 1) Component-Level Ablation

We conduct a detailed ablation study on the M3D dataset to evaluate the contributions of five key components: the original ConvNeXt local feature branch, the ViT-based global contextual branch, the High-Frequency Enhancement Module (HFM), the ERP-aware padding mechanism (ERP Pad.), and the Channel Attention (CA) module. The combined results for both tasks, camera inclination angle estimation and upright panorama generation, are summarized in Table 7.

The first row illustrates the configuration using only ViT-derived global contextual features. Its relatively low inclination-estimation accuracy indicates that the ViT-only representation is less effective for this task than configurations incorporating the ERP-based CNN branch. For panorama generation, it also yields lower image quality, highlighting the importance of local structural features.

Rows #2–#5 focus on the ConvNeXt-based local branch. As ERP-aware padding, HFM, and CA are incrementally introduced, inclination-estimation accuracy and image-generation quality generally improve. These results confirm the complementary contributions of boundary-aware processing, high-frequency structural enhancement, and channel-wise refinement.

Finally, Row #6 introduces the ViT branch and dual-projection fusion. The strict-threshold angle accuracy changes only slightly compared with the best local-only configuration, whereas image-generation quality improves noticeably, with FID decreasing from 3.95 to 3.26 and NRMSE from 0.1809 to 0.1728. This is consistent with the network design, since the ViT-derived fused features are primarily used by the panorama-generation decoder, while inclination estimation directly operates on the deepest ERP feature.

Table 10.
ROBUSTNESS ANALYSIS OF THE PROPOSED MODEL UNDER SIMULATED DEGRADATION - MOSAIC OCCLUSION

| Mosaic Occlusion | | Accuracy within Angular-Error Thresholds (%). | | | | | | Quality of image generation | | |
|---|---|---|---|---|---|---|---|---|---|---|
| | | <1° | <2° | <3° | <4° | <5° | <12° | FID↓ | NRMSE↓ | NMAE↓ |
| Occlusion Mask Sizes | **128** | 15.8 | 33.4 | 46.6 | 57.9 | 67.4 | 95.6 | 6.63 | 0.2387 | 0.1581 |
| | **96** | 31.0 | 58.4 | 73.4 | 82.8 | 88.4 | 99.2 | 6.12 | 0.2133 | 0.1383 |
| | **64** | 49.6 | 82.7 | 93.1 | 97.2 | 98.6 | 99.8 | 5.92 | 0.1956 | 0.1252 |
| | **32** | 63.2 | 94.1 | 98.6 | 99.5 | 99.8 | 100 | 5.87 | 0.1876 | 0.1194 |
| | / | 65.9 | 95.5 | 99.0 | 99.7 | 99.9 | 100 | 5.87 | 0.1859 | 0.1182 |

Table 11.
ROBUSTNESS ANALYSIS OF THE PROPOSED MODEL UNDER SIMULATED DEGRADATION - REALISTIC OCCLUSION

| Realistic Mask-based Occlusion | | Accuracy within Angular-Error Thresholds (%). | | | | | | Quality of image generation | | |
|---|---|---|---|---|---|---|---|---|---|---|
| | | <1° | <2° | <3° | <4° | <5° | <12° | FID↓ | NRMSE↓ | NMAE↓ |
| Occlusion Field of View (FOV) | **80°** | 42.2 | 77.8 | 91.0 | 95.9 | 97.8 | 99.8 | 6.00 | 0.2000 | 0.1283 |
| | **60°** | 51.5 | 86.8 | 95.8 | 98.4 | 99.3 | 100 | 5.94 | 0.1936 | 0.1237 |
| | **40°** | 59.5 | 92.8 | 98.1 | 99.2 | 99.6 | 100 | 5.88 | 0.1891 | 0.1204 |
| | **20°** | 65.0 | 94.7 | 98.8 | 99.6 | 99.9 | 100 | 5.86 | 0.1865 | 0.1186 |
| | / | 65.9 | 95.5 | 99.0 | 99.7 | 99.9 | 100 | 5.87 | 0.1859 | 0.1182 |

### 2) Multi-task Collaboration Analysis

To examine the effect of collaborative multi-task learning, we compare an angle-only model, an image-generation-only model, and the full multi-task model.

As shown in Table 8, joint training improves the 1° inclination accuracy from 81.9% to 85.2%, indicating that generation supervision benefits the shared representation used for inclination estimation. For panorama generation, FID decreases from 3.68 to 3.26, whereas NRMSE and NMAE increase slightly. This suggests that inclination-angle supervision introduces complementary orientation-sensitive constraints into the shared representation, improving perceptual quality without uniformly improving all pixel-level metrics.

### 3) Projection Design Validation

To validate the dual-projection design, we compare the proposed configuration (ERP for the CNN branch and cubemap for the ViT branch) with a single-projection variant in which both branches use ERP inputs, while keeping the remaining training settings unchanged.

As shown in Table 9, the proposed ERP + CMP configuration consistently improves inclination-estimation accuracy over the ERP + ERP variant, particularly under strict error thresholds (85.2% vs. 83.7% at <1° and 95.0% vs. 94.7% at <2°). It also achieves slightly better upright panorama generation quality, reducing FID from 3.42 to 3.26, NRMSE from 0.1733 to 0.1728, and NMAE from 0.1129 to 0.1128. These results support the use of complementary projection representations: ERP preserves continuous panoramic structure for CNN-based local modeling, while cubemap reduces latitude-dependent distortion and provides more uniform patches for ViT-based global contextual modeling.

## *D. Robustness Analysis*

To evaluate the robustness of orientation estimation and spherical mapping prediction under adverse visual conditions, we simulate three controlled degradations, mosaic occlusion, realistic object-based occlusion, and Gaussian noise, on the SUN360 test set. Inclination estimation robustness is evaluated directly using angular errors. To isolate the effect of input corruption on the predicted geometric correction itself, we adopt a decoupled evaluation protocol: the model predicts the inclination angles and 3D spherical LUT from the corrupted non-upright panorama, while the predicted LUT is applied to the corresponding clean non-upright panorama for resampling. The reconstructed upright panorama is then compared with the clean ground truth using FID, NRMSE, and NMAE. Under this protocol, changes in reconstruction quality primarily reflect errors in the predicted spherical mapping rather than visual degradation already present in the corrupted input. Therefore, this experiment should be interpreted as a controlled diagnostic evaluation of correction robustness, rather than as an evaluation of image restoration from degraded inputs.

**Mosaic Occlusion Evaluation:** Mosaic occlusion is adopted as a controlled perturbation to systematically evaluate the robustness of the proposed framework. Compared with realistic object occlusion, it provides precise control over the occlusion extent and ensures reproducible evaluation conditions. At the same time, mosaic masking removes local structural cues important for inclination estimation and introduces artificial block boundaries that are inconsistent with the underlying panoramic geometry, thereby creating a challenging test for methods relying on geometric information. We apply random square masks of sizes {32×32, 64×64, 96×96, 128×128} to non-upright 256×512 ERP inputs using a fixed mosaic block size of 10 pixels.

As shown in Table 10, performance gradually degrades as the occlusion size increases. With a 64×64 mask, angle accuracy within 3° remains 93.1%, and FID is 5.92, close to the unobstructed baseline of 5.87. Under the most severe 128×128 occlusion, Acc@1° drops to 15.8% and FID increases to 6.63. Under the decoupled protocol, these image-quality changes reflect the effect of mosaic corruption on spherical mapping prediction rather than the direct visual appearance of the occlusion itself. The stronger degradation under severe masking suggests that missing local structures and artificial block boundaries increasingly interfere with both inclination estimation and spherical mapping prediction.

**Table 12.**
ROBUSTNESS ANALYSIS OF THE PROPOSED MODEL UNDER SIMULATED DEGRADATION - GAUSSIAN NOISE

| Gaussian Noise | | Accuracy within Angular-Error Thresholds (%). | | | | | | Quality of image generation | | |
|---|---|---|---|---|---|---|---|---|---|---|
| | | <1° | <2° | <3° | <4° | <5° | <12° | FID↓ | NRMSE↓ | NMAE↓ |
| Noise Levels (Sigma) | **0.05** | 37.2 | 74.0 | 88.7 | 94.3 | 96.5 | 99.3 | 6.28 | 0.2097 | 0.1365 |
| | **0.04** | 47.5 | 83.6 | 94.5 | 97.2 | 98.5 | 99.7 | 6.05 | 0.1984 | 0.1276 |
| | **0.03** | 56.9 | 90.2 | 97.1 | 98.9 | 99.5 | 99.9 | 5.93 | 0.1918 | 0.1228 |
| | **0.02** | 61.4 | 93.2 | 98.3 | 99.4 | 99.8 | 100 | 5.90 | 0.1883 | 0.1201 |
| | **0.01** | 64.7 | 95.0 | 98.8 | 99.6 | 99.9 | 100 | 5.83 | 0.1865 | 0.1187 |
| | / | 65.9 | 95.5 | 99.0 | 99.7 | 99.9 | 100 | 5.87 | 0.1859 | 0.1182 |

**Table 13.**
ROBUSTNESS ANALYSIS OF THE NOISE-AWARE FINE-TUNED MODEL UNDER SIMULATED GAUSSIAN NOISE

| Gaussian Noise | | Accuracy within Angular-Error Thresholds (%). | | | | | | Quality of image generation | | |
|---|---|---|---|---|---|---|---|---|---|---|
| | | <1° | <2° | <3° | <4° | <5° | <12° | FID↓ | NRMSE↓ | NMAE↓ |
| Noise Levels (Sigma) | **0.05** | 51.9 | 86.6 | 95.8 | 98.3 | 99.1 | 99.9 | 6.18 | 0.1920 | 0.1230 |
| | **0.04** | 59.9 | 91.9 | 97.6 | 99.1 | 99.6 | 99.9 | 6.15 | 0.1867 | 0.1193 |
| | **0.03** | 67.0 | 94.9 | 98.6 | 99.4 | 99.8 | 99.9 | 6.14 | 0.1833 | 0.1169 |
| | **0.02** | 68.7 | 95.7 | 98.9 | 99.6 | 99.8 | 100 | 6.13 | 0.1819 | 0.1158 |
| | **0.01** | 69.6 | 96.2 | 99.0 | 99.6 | 99.8 | 100 | 6.13 | 0.1816 | 0.1156 |
| | / | 69.9 | 96.2 | 99.1 | 99.7 | 99.8 | 100 | 6.13 | 0.1815 | 0.1155 |

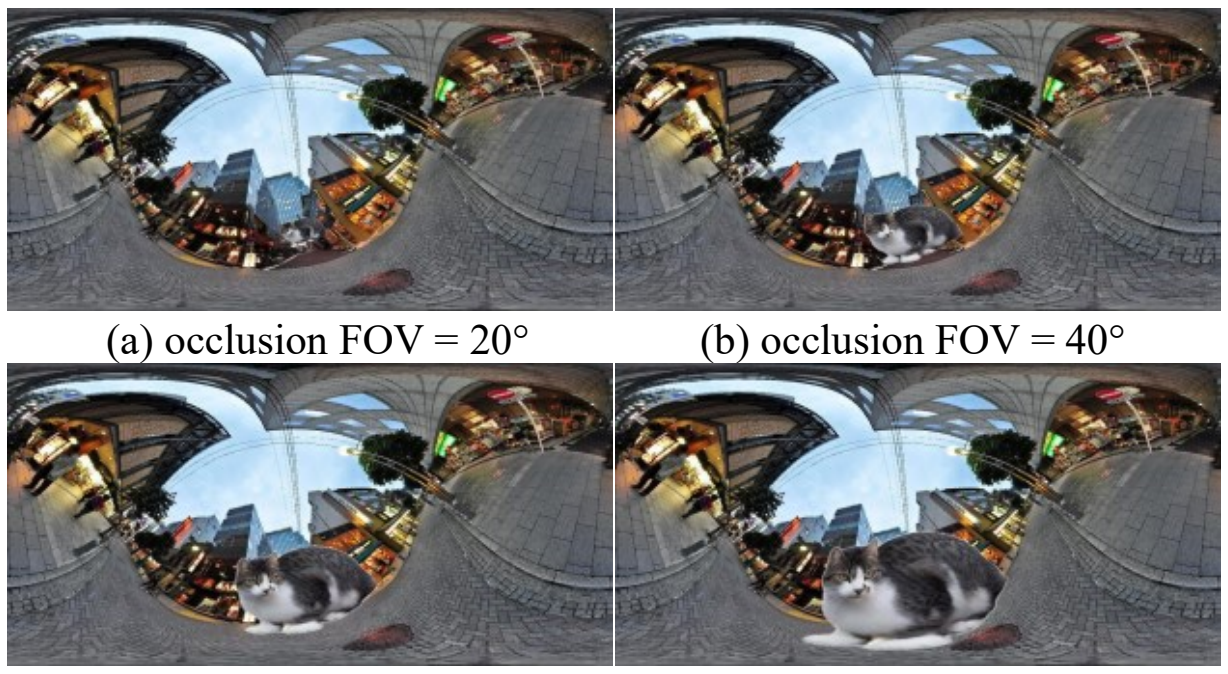

(a) occlusion FOV = 20° (b) occlusion FOV = 40°
(c) occlusion FOV = 60° (d) occlusion FOV = 80°

**Fig. 9.** Examples of realistic object-based occlusion under different FOV settings on a non-upright ERP image.

**Realistic Occlusion Evaluation:** To assess robustness under more practical occlusion patterns, segmented real-world objects are projected onto the spherical domain and fused with the spherical representation of the upright panorama. For each test sample, a spherical location is randomly sampled using a longitude–latitude pair, and the object is placed at the corresponding position on the sphere. For the same test image, this location is kept fixed across different occlusion levels, while the severity is controlled by varying the object field of view (FOV ∈ {20°, 40°, 60°, 80°}), as illustrated in Fig. 9. The occluded spherical panorama is then converted back to ERP and rotated using the same ground-truth inclination as the original sample, ensuring a consistent simulated camera orientation across evaluations. Unlike mosaic masking, this strategy preserves natural object shapes and texture continuity and therefore provides a more realistic test of partial scene occlusion.

As shown in Table 11, the proposed method remains relatively robust under realistic occlusion. At FOV = 60°, angle accuracy within 3° reaches 95.8%, while at FOV = 80°, Acc@1° remains 42.2%, substantially higher than the 15.8% observed under the strongest mosaic occlusion. FID increases only from 5.87 without occlusion to 6.00 at FOV = 80°. Under the decoupled protocol, the limited variation in reconstruction metrics indicates that realistic occlusion has a relatively small effect on spherical mapping prediction. The smaller degradation compared with mosaic masking may be attributed to the absence of artificial block boundaries and the preservation of more natural local structures.

**Gaussian Noise Corruption:** Gaussian noise is introduced to evaluate sensitivity to low-level appearance corruption that perturbs local intensity and edge information without explicitly removing image regions. We add zero-mean Gaussian noise with standard deviations $\sigma \in \{0.01, 0.02, 0.03, 0.04, 0.05\}$ to the input images. As shown in Table 12, performance remains relatively stable under low noise levels but gradually degrades as the noise increases. At $\sigma = 0.05$, Acc@1° decreases from 65.9% to 37.2%, while FID increases from 5.87 to 6.28. These results indicate that Gaussian noise affects both inclination estimation and spherical mapping prediction, with inclination estimation showing greater sensitivity under strict angular error thresholds. Under the decoupled protocol, the comparatively smaller changes in FID, NRMSE, and NMAE indicate that spherical mapping prediction degrades more gradually than inclination estimation as the noise level increases.

**Noise-aware Training:** To further improve robustness under Gaussian noise, we investigate a noise-aware fine-tuning strategy using noise-augmented training data. Gaussian noise with $\sigma$ uniformly sampled from 0 to 0.05 is injected into the training images, and the pretrained model is fine-tuned for 20 epochs using a learning rate of $1\times10^{-5}$. Compared with Table 12, Table 13 shows improved inclination-estimation accuracy and lower NRMSE and NMAE across the tested noise levels. At $\sigma$=0.05, Acc@1° increases from 37.2% to 51.9%, while NRMSE decreases from 0.2097 to 0.1920. Although FID does not improve uniformly, it decreases from 6.28 to 6.18 under the strongest corruption. A slight degradation in clean-input FID is also observed after fine-tuning (5.87 to 6.13), indicating a modest trade-off between noise robustness and clean-data perceptual quality. Overall, these results demonstrate that noise-aware fine-tuning can mitigate performance degradation under Gaussian noise, although the magnitude of improvement varies across metrics. Although this experiment focuses on Gaussian noise corruption, similar robustness-oriented training strategies could potentially be extended to structured occlusions such as mosaic masking.

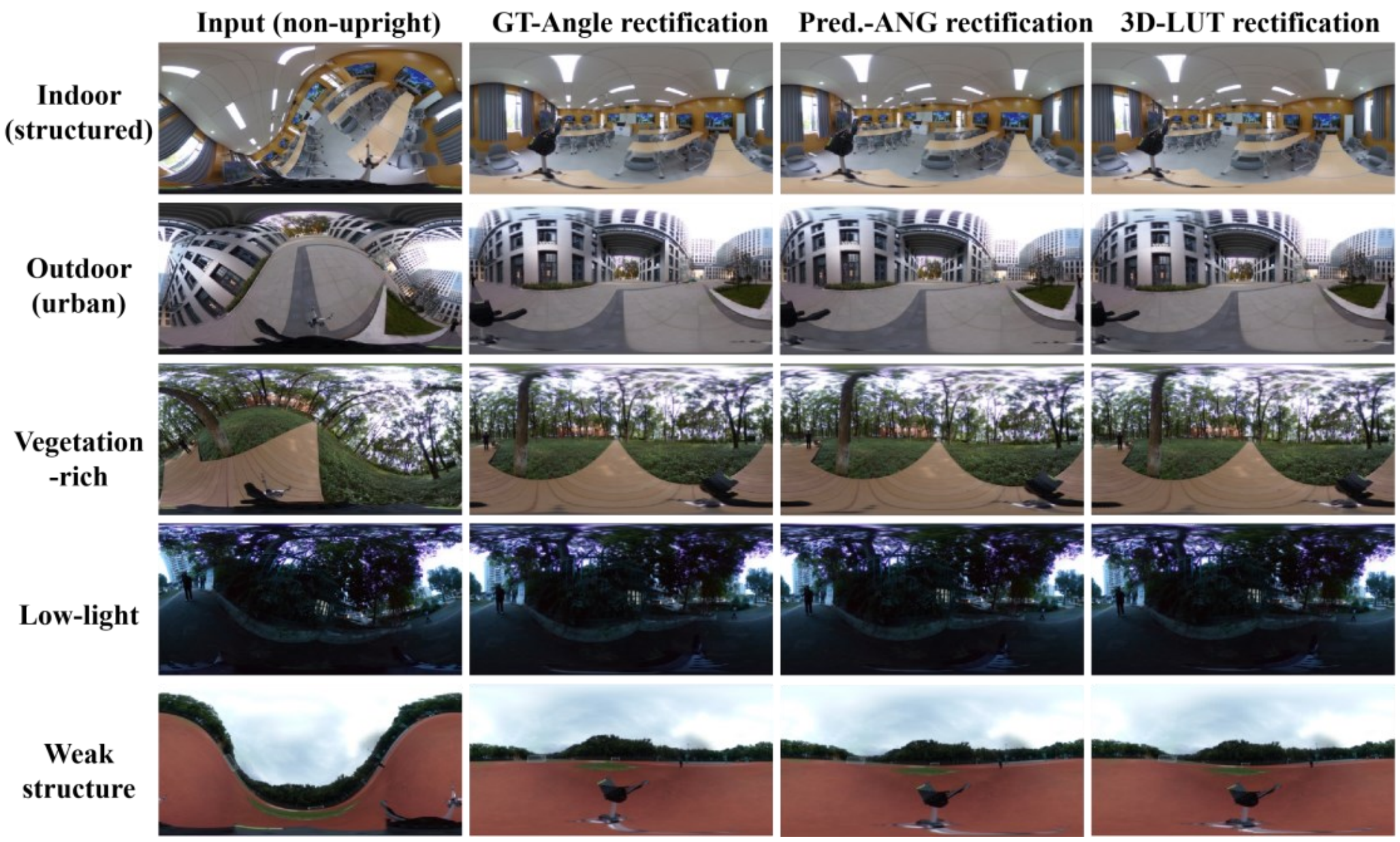

**Fig. 10. Representative qualitative rectification results on real-world RICOH THETA panoramas.** From left to right: non-upright input, geometric rectification using the ground-truth orientation, geometric rectification using the predicted orientation, and learned 3D spherical-LUT rectification. The examples include indoor, outdoor, vegetation-rich, low-light, and open-scene conditions.

**Table 14.**
REAL-WORLD INCLINATION-ESTIMATION PERFORMANCE OF THE ORIGINAL AND NOISE-AWARE MODELS

| Scene | Model | Angular-error (°) | | | | Accuracy within Angular-Error Thresholds (%). | | | | | |
|---|---|---|---|---|---|---|---|---|---|---|---|
| | | MAE ↓ | Median ↓ | RMSE ↓ | P95 ↓ | <1°↑ | <3°↑ | <5°↑ | <7°↑ | <10°↑ | <12°↑ |
| **Overall (200)** | **Original** | 11.36 | 7.84 | 15.96 | 39.28 | 1.50 | 14.50 | 27.00 | 46.00 | 64.00 | 69.50 |
| | **Noise-Aware** | 7.38 | 4.66 | 14.86 | 22.79 | 3.00 | 27.50 | 54.50 | 79.00 | 91.50 | 93.50 |
| **Indoor (68)** | **Original** | 8.22 | 4.90 | 13.20 | 30.84 | 4.41 | 25.00 | 50.00 | 69.12 | 85.29 | 85.29 |
| | **Noise-Aware** | 7.75 | 3.29 | 18.11 | 40.35 | 2.94 | 47.06 | 72.06 | 89.71 | 91.18 | 91.18 |
| **Outdoor (71)** | **Original** | 12.15 | 9.51 | 16.63 | 39.22 | 0.00 | 11.27 | 16.90 | 42.25 | 54.93 | 63.38 |
| | **Noise-Aware** | 8.15 | 4.91 | 16.31 | 24.93 | 2.82 | 21.13 | 50.70 | 77.46 | 90.14 | 94.37 |
| **Vegetation rich (61)** | **Original** | 13.94 | 9.60 | 17.85 | 39.26 | 0.00 | 6.56 | 13.11 | 24.59 | 50.82 | 59.02 |
| | **Noise-Aware** | 6.06 | 6.21 | 6.95 | 10.00 | 3.28 | 13.11 | 39.34 | 68.85 | 93.44 | 95.08 |

The decoupled protocol does not reproduce the complete degraded-input-to-output deployment pipeline; rather, it is designed to isolate the robustness of geometric correction prediction. Real-world camera evaluation is therefore considered separately in Sec. IV-E.

### *E. Real-World Evaluation and Failure-Case Analysis*

#### 1) Real-World Rectification Evaluation

To evaluate synthetic-to-real generalization, we constructed a 200-image real-world panoramic evaluation set using a RICOH THETA camera, including 68 indoor, 71 outdoor, and 61 vegetation-rich scenes. Automatic top–bottom correction was disabled during acquisition, and pitch/roll annotations were obtained from the camera-recorded orientation metadata. The acquisition was designed to cover **challenging conditions**, including a wide range of camera orientations (pitch: −53.10° to 60.71°; roll: −75.93° to 87.09°), illumination ranging from low-light environments to strong outdoor daylight, structured and weakly structured scenes, occlusion, narrow spaces, and large low-texture regions such as white walls. This set is used as **an additional real-world benchmark** and has been released through our GitHub repository together with the annotations and evaluation scripts.

#### 2) Quantitative Uprightness Evaluation

Table 14 reports the inclination-estimation errors and threshold-based accuracies on the real-world set. All experiments are conducted using the SUN360-trained model, as SUN360 contains more diverse indoor and outdoor scene content than M3D. Compared with the original model, Gaussian-noise-aware training reduces the MAE from 11.36° to 7.38° and the median error from 7.84° to 4.66°, while Acc@5° and Acc@10° increase from 27.0% and 64.0% to 54.5% and 91.5%, respectively. The improvement is especially evident in outdoor and vegetation-rich scenes. For vegetation-rich scenes, the MAE decreases from 13.94° to 6.06° and P95 from 39.26° to 10.00°. These results indicate that noise-aware training improves robustness to the real-camera domain shift, although difficult cases remain.

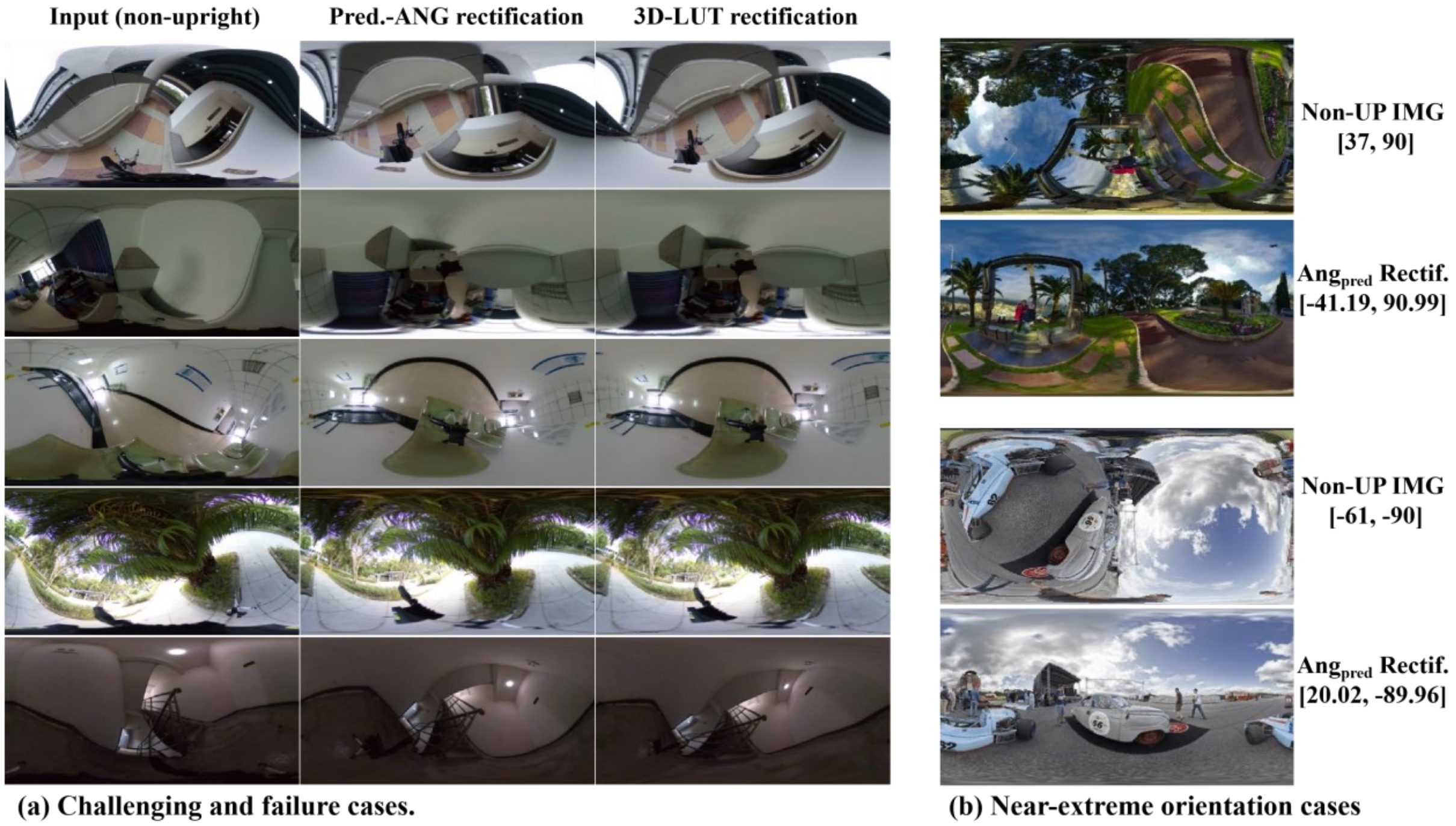

**Fig. 11. Representative challenging and failure cases from the controlled benchmarks and the real-world THETA set.** (a) Real-world challenging and failure cases under combinations of adverse factors, including large inclination, close-range scene geometry, misleading local structures, repetitive textures, and low illumination. (b) Near-extreme orientation cases in which large pitch–roll discrepancies can still result in visually similar upright rectification.

The lower accuracy on real-world panoramas than on the synthetic SUN360 and M3D benchmarks reflects the substantial difference between the two evaluation settings. Although SUN360 and M3D contain diverse real-world scene content, their non-upright benchmark inputs are generated through controlled spherical rotation. In contrast, real THETA acquisition involves a complete dual-fisheye imaging and stitching pipeline and introduces additional camera-specific effects, such as stitching artifacts, exposure variation, and sensor noise, which are not fully reproduced by synthetic rotation. Moreover, foliage, weak structural cues, repetitive textures, low illumination, and stitching-boundary artifacts can further reduce the reliability of orientation estimation. These results reveal a clear synthetic-to-real domain gap rather than simply an effect of inclination magnitude.

#### 3) Qualitative Rectification and Failure-Case Analysis

Representative real-world rectification results are shown in Fig. 10, including the original non-upright panorama, the result rectified using the ground-truth orientation, the predicted-angle rectification, and the learned 3D spherical-LUT rectification. The successful examples show that both rectification pathways can recover the dominant upright scene structure under diverse real-world conditions, including indoor, outdoor, vegetation-rich, low-light, and open scenes, providing complementary qualitative evidence to the quantitative evaluation above.

Fig. 11 further presents representative challenging and failure cases from both the controlled benchmarks and the real-world THETA set. The observed failures are generally associated with combinations of adverse factors rather than a single cause, including large camera inclination, close-range scene geometry, misleading dominant local structures, repetitive textures, and stitching or projection artifacts. In some cases, strong but unreliable orientation cues may dominate the prediction, such as horizontal structures being interpreted as vertical cues or repetitive ceiling patterns introducing conflicting directional information. Fig. 11 also presents near-extreme orientation cases in which the ground-truth and predicted pitch–roll values differ substantially while their rectified panoramas remain visually similar. This phenomenon is particularly evident when the roll component approaches ±90° under the adopted rotation convention, where the pitch–roll representation becomes highly sensitive and large component-wise differences do not necessarily correspond to equally large differences in upright orientation.

Overall, the real-world evaluation demonstrates the applicability of the proposed method to challenging real-camera imagery while revealing a remaining synthetic-to-real gap. Future work will investigate real-image adaptation and domain-adaptation strategies to further improve robustness.

### *F. Runtime and Deployment Analysis*

To evaluate the practical applicability of the proposed method, we analyze the runtime performance and computational complexity of different model variants on a workstation equipped with an Intel Core i9-11900K CPU, 64 GB of DDR4-3200 RAM, and an NVIDIA RTX 3090 GPU with 24 GB memory. Specifically, we report the number of floating-point operations (FLOPs), model parameters, model size, inference latency, and corresponding frames per second (FPS). All measurements are performed with a batch size of 1 and an input resolution of 256×512. Before timing, 50 warm-up iterations are performed, followed by 200 timed runs measured using CUDA Events. Latency is reported as the mean ± standard deviation over the 200 timed runs, and FPS is calculated from the mean latency. The reported latency covers only the network forward pass and excludes disk I/O, ERP-to-cubemap preprocessing, and post-forward LUT conversion and image resampling. The comparison includes three configurations: an angle-only deployment configuration, the full model with implicit data-driven cross-projection transformation, and the full model with explicit geometric reprojection, as summarized in Table 15.

**Table 15.**
RUNTIME PERFORMANCE AND MODEL COMPLEXITY ON RTX 3090

| Variant | FLOPs (G) | Params (M) | Size (MB) | Latency (ms) | FPS |
|---|---|---|---|---|---|
| **Angle-Only** (from Full Imp-Trans) | 40.11 | 87.51 | 338 | 14.97±0.70 | 66.82 |
| **Full Model** (Imp-Trans) | 114.80 | 246.71 | 946 | 36.83±1.96 | 27.15 |
| **Full Model** (Exp-Proj) | 112.36 | 240.46 | 971 | 65.90±5.44 | 15.17 |

The **angle-only** deployment configuration is extracted from the jointly trained full Imp-Trans model and retains only the ERP ConvNeXt encoder and the angle-regression head, while bypassing the ViT/cubemap branch, cross-projection fusion modules, and LUT-generation decoder during inference. Importantly, this reduced configuration produces exactly the same inclination predictions as the full Imp-Trans model, while reducing the computational cost to 40.11 GFLOPs and 87.51 M parameters. It achieves an average latency of 14.97 ± 0.70 ms, corresponding to 66.82 FPS.

The full model with implicit data-driven feature transformation jointly performs inclination estimation and upright panorama generation with a latency of 36.83 ± 1.96 ms per frame, corresponding to 27.15 FPS. In comparison, the explicit geometric projection variant requires 65.90 ± 5.44 ms and achieves 15.17 FPS. Thus, the implicit transformation provides substantially lower computational overhead while maintaining comparable rectification performance.

These results demonstrate that different deployment configurations can be selected according to application requirements. When only camera inclination estimation is required, the angle-only configuration provides the highest throughput at approximately 67 FPS while preserving the angle predictions of the full Imp-Trans model. When both inclination estimation and upright panorama generation are required, the full Imp-Trans model operates at approximately 27 FPS and remains substantially faster than the explicit-projection variant.

To further evaluate resolution-independent rectification, we apply the inclination estimates produced by the angle-only configuration to higher-resolution panoramas. The angle-estimation branch continues to operate on 256×512 ERP inputs, exactly as during training, while the predicted pitch and roll are applied through deterministic spherical resampling to the corresponding 512×1024 panoramas. Thus, the higher-resolution panoramas are used only in the geometric rectification stage and are not forwarded through the neural network, requiring neither retraining nor high-resolution network inference.

For the geometric rectification stage, the resolution-dependent spherical base grid is precomputed once and excluded from the per-frame runtime measurement. The reported rectification latency includes rotation-matrix construction, spherical-coordinate transformation, ERP sampling-grid generation, and image resampling using ***grid_sample***. The same timing protocol of 50 warm-up runs followed by 200 CUDA-Event measurements is adopted. As shown in **Table 16**, geometric rectification requires only 0.67 ± 0.03 ms at 256×512 and 0.76 ± 0.09 ms at 512×1024. Combined with the 14.97 ± 0.70 ms angle-estimation latency, the complete angle-estimation-and-rectification pipeline requires approximately 15.64 ms and 15.73 ms per frame for 256×512 and 512×1024 outputs, respectively, corresponding to 63.96 FPS and 63.57 FPS. The small increase at 512×1024 confirms that higher output resolution introduces only limited geometric-resampling overhead because neural-network inference remains fixed at 256×512.

**Table 16.**
RUNTIME PERFORMANCE OF ANGLE-BASED GEOMETRIC RECTIFICATION AT DIFFERENT RESOLUTIONS

| Rectification resolution | Geometric rectification latency (ms) | Total pipeline latency (ms) | FPS |
|---|---|---|---|
| **256×512** | 0.67 ± 0.03 | 15.64 ± 0.70 | 63.96 |
| **512×1024** | 0.76 ± 0.09 | 15.73 ± 0.71 | 63.57 |

For deployment on embedded platforms such as NVIDIA Jetson devices, the proposed architecture mainly consists of standard convolutional, attention, and transformer operations that are compatible with modern edge inference frameworks such as TensorRT. Although embedded-platform runtime is not directly evaluated in the present study, a practical deployment path includes mixed-precision inference (FP16), INT8 quantization, lightweight CNN/ViT backbones, model pruning, and knowledge distillation. Recent lightweight optimization methods, including self-knowledge distillation [42], may provide additional efficiency gains. The angle-only configuration further provides a directly validated reduced-computation mode, reducing latency from 36.83 ± 1.96 ms to 14.97 ± 0.70 ms while preserving exactly the same inclination predictions. Future work will evaluate these optimization strategies on resource-constrained platforms such as Jetson Orin or Xavier.

## V. Conclusions and Future Work

In this paper, we present a projection-aware dual-projection framework for vision-based upright rectification of 360° panoramic images. The proposed method jointly performs inclination estimation and upright panorama generation by exploiting complementary representations from two projection domains. Specifically, a CNN branch captures local structural cues from ERP inputs, while a ViT branch models global contextual information from cubemap projections. Cross-projection feature transformation and multi-level fusion enable effective interaction between these complementary representations. In addition, high-frequency enhancement and circular padding are introduced to better preserve structural cues and panoramic continuity. Extensive experiments on SUN360 and M3D demonstrate the effectiveness of the proposed method in both inclination estimation and upright panorama generation, while ablation studies verify the contributions of dual-projection representation learning, cross-projection interaction, panoramic-specific structural modules, and collaborative multi-task learning. Additional real-world experiments further evaluate the generalization of the method beyond synthetically rotated benchmark images.

Despite its competitive performance, several limitations remain. The model is still sensitive to degradations that substantially weaken or distort reliable orientation cues, such as large-area occlusion, strong noise, repetitive or misleading local structures, and complex close-range geometry. This sensitivity is mainly reflected in inclination estimation accuracy, while the predicted spherical mapping generally degrades more gradually under the controlled perturbations evaluated using the decoupled robustness protocol. The noise-aware experiment suggests that targeted degradation augmentation is a promising direction for improving robustness. Future work may therefore investigate more general degradation-aware training and robustness-oriented representation learning. In addition, the distribution gap between synthetically rotated benchmark panoramas and real-camera imagery remains an important issue. Environment-aware or domain-adaptation strategies [43] may therefore provide a promising direction for improving cross-domain generalization.

Moreover, the learned generation branch currently operates at a fixed resolution of 256×512, which is mainly adopted for computational efficiency and fair comparison with existing methods. Since the estimated inclination angles describe camera orientation rather than image resolution, they can also be directly applied to higher-resolution panoramas through standard geometric resampling. Extending the generation framework to higher resolutions through hierarchical decoding or multi-resolution processing remains an important direction. The dual-encoder architecture also introduces non-negligible computational cost, motivating future investigation of lightweight backbones, compact transformers, pruning, distillation, quantization, and more efficient cross-projection transformation mechanisms for resource-constrained platforms.

Further work may also investigate more fine-grained scenario-specific analysis to better understand the respective roles of global and local cues in panoramic rectification. Recent neural representations for flexible 3D surface modeling and unconstrained viewpoint rendering [44] also suggest broader opportunities for extending panoramic orientation correction toward 3D visual representation and rendering pipelines. Visual–inertial integration represents another potential extension, where the proposed visual orientation estimates could complement inertial measurements for consistency checking and redundancy under sensor degradation. Addressing these issues may further improve the robustness and applicability of the proposed framework in real-world panoramic vision systems.

**Acknowledgements:** This research was funded by the Fundamental Research Funds for the Central Universities (Grant No. SWU120083), Science and Technology Innovation Key R&D Program of Chongqing (CSTB2023TIAD-STX0037), the National Natural Science Foundation of China (Grant No. 6267340).

**Author Contributions:** Yuhao Shan: Conceptualization, Methodology, Validation, Formal Analysis, and Writing - Original Draft. Qianyi Yuan: Writing - Original Draft (language editing and revision). Jingguo Liu: Conceptualization. Shigang Li: Conceptualization. Jianfeng Li: Conceptualization, Review & Editing (manuscript structure and logic optimization). Tong Chen: Writing - Review & Editing (logic optimization, language polishing and final proofreading).

**Data Availability:** The code and related datasets can be found at: https://github.com/YuhaoShine/DualProjectionFusion.

**Declarations - Conflict of interest:** All authors certify that they have no affiliations with or involvement in any organization or entity with any financial interest or non-financial interest in the subject matter or materials discussed in this manuscript.

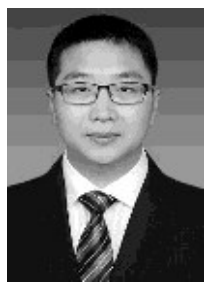

**Yuhao Shan** received the M.S. degree in signal and information processing from Southwest University, Chongqing, China, in 2016. He received the Ph.D. degree in information science from Hiroshima City University, Hiroshima, Japan, in 2020. He is currently a Lecturer with Southwest University. His research interests include computer vision, signal processing, and affective computing.

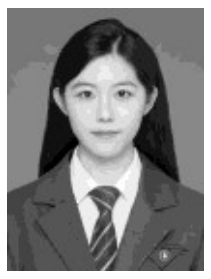

**Qianyi Yuan** received the B.S. degree in information security from the College of Electronic and Information Engineering, Southwest University, Chongqing, China, in 2026. Her research interests include computer vision, signal processing, and pattern recognition.

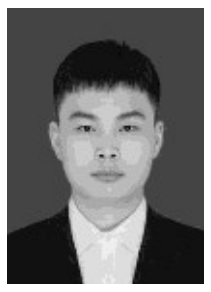

**Jingguo Liu** received the M.S. degree in information and communication engineering from Southwest University in 2025, as part of an integrated undergraduate–graduate program. His research interests include computer vision, multimodal large language models, and spatial reasoning.

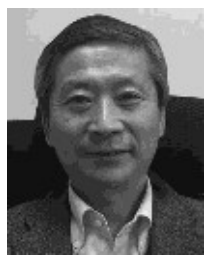

**Shigang Li** received the M.S. and Ph.D. degrees in control engineering from Osaka University, Toyonaka, Japan, in 1990 and 1993, respectively. He is currently a Professor of Hiroshima City University, Hiroshima, Japan. His research interests include computer/robot vision, human–machine interface, and intelligent robot.

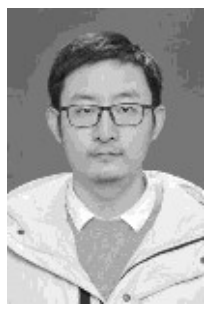

**Jianfeng Li** received the M.S. degree from the College of Computer Science, Sichuan University, China, in 2013, and the Ph.D. degree in information engineering from Tottori University, Japan, in 2017. He is currently a Lecturer at Southwest University, Chongqing, China. His primary research interest focuses on the field of computer vision with in-depth research in related technical applications.

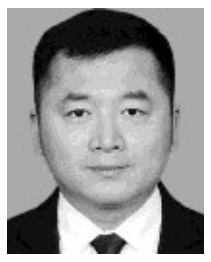

**Tong Chen** received the M.S. degree in digital sound and vision processing from the University of Wales, Wales, U.K., in 2006, and the Ph.D. degree information science from Cranfield University, Cranfield, U.K., in 2013. He is currently a Professor with Southwest University, Chongqing, China. His research interests include affective computing, neural networks, and computational intelligence.